\documentclass[10pt,twocolumn,letterpaper]{article}

\usepackage[pagenumbers]{cvpr} % To force page numbers, e.g. for an arXiv version

\usepackage{graphicx}
\usepackage{amsmath}
\usepackage{amssymb}
\usepackage[accsupp]{axessibility}  % Improves PDF readability for those with disabilities.
\usepackage{booktabs}
\usepackage{color}
\definecolor{pink}{RGB}{255, 192, 203}
\definecolor{gold}{RGB}{255, 192, 0}
\definecolor{purple}{RGB}{145, 12, 195}
\usepackage{multirow}
\newcounter{RNum}
\renewcommand{\theRNum}{\arabic{RNum}}
\newcommand{\Remark}{\noindent\textit{\textbf{Remark}~\refstepcounter{RNum}\textbf{\theRNum}: }}
\newcommand{\NoOne}[1]{\textcolor{red}{#1}}
\newcommand{\NoTwo}[1]{\textcolor{green}{#1}}
\newcommand{\NoThree}[1]{\textcolor{blue}{#1}}
\newcommand{\NoFour}[1]{\textcolor{gold}{#1}}
\newcommand{\pink}[1]{\textcolor{pink}{#1}}
\newcommand{\purple}[1]{\textcolor{purple}{#1}}
\usepackage{soul}
\soulregister\NoOne7
\soulregister\NoTwo7
\soulregister\NoThree7

\usepackage[pagebackref,breaklinks,colorlinks]{hyperref}

\usepackage[capitalize]{cleveref}
\crefname{section}{Sec.}{Secs.}
\Crefname{section}{Section}{Sections}
\Crefname{table}{Table}{Tables}
\crefname{table}{Tab.}{Tabs.}

\def\cvprPaperID{2683} % *** Enter the CVPR Paper ID here
\def\confName{CVPR}
\def\confYear{2022}

\begin{document}
	
	%%%%%%%%% TITLE - PLEASE UPDATE
	\title{Unsupervised Domain Adaptation for Nighttime Aerial Tracking}
	
	\author{Junjie Ye$^{\dagger}$, Changhong Fu$^{\dagger,}$\thanks{Corresponding author}~, Guangze Zheng$^{\dagger}$, Danda Pani Paudel$^{\ddagger}$, and Guang Chen$^{\dagger}$ \\
		$^{\dagger}$Tongji University, China \quad $^{\ddagger}$ETH Zürich, Switzerland\\
		{\tt\small \{ye.jun.jie, changhongfu, mmlp, guangchen\}@tongji.edu.cn, paudel@vision.ee.ethz.ch}
		% For a paper whose authors are all at the same institution,
		% omit the following lines up until the closing ``}''.
		% Additional authors and addresses can be added with ``\and'',
		% just like the second author.
		% To save space, use either the email address or home page, not both
%		\and
%		Second Author\\
%		Institution2\\
%		First line of institution2 address\\
%		{\tt\small secondauthor@i2.org}
	% 
	}
	\maketitle

	%%%%%%%%% ABSTRACT
	\begin{abstract}
		Previous advances in object tracking mostly reported on favorable illumination circumstances while neglecting performance at nighttime, which significantly impeded the development of related aerial robot applications. This work instead develops a novel unsupervised domain adaptation framework for nighttime aerial tracking (named UDAT). Specifically, a unique object discovery approach is provided to generate training patches from raw nighttime tracking videos. To tackle the domain discrepancy, we employ a Transformer-based bridging layer post to the feature extractor to align image features from both domains. With a Transformer day/night feature discriminator, the daytime tracking model is adversarially trained to track at night. Moreover, we construct a pioneering benchmark namely NAT2021 for unsupervised domain adaptive nighttime tracking, which comprises a test set of 180 manually annotated tracking sequences and a train set of over 276k unlabelled nighttime tracking frames. Exhaustive experiments demonstrate the robustness and domain adaptability of the proposed framework in nighttime aerial tracking. The code and benchmark are available at \url{https://github.com/vision4robotics/UDAT}.
	\end{abstract}
	
	%%%%%%%%% BODY TEXT
	\section{Introduction}
	\label{sec:intro}
	Standing as one of the fundamental tasks in computer vision, object tracking has received widespread attention with multifarious aerial robot applications, \eg, unmanned aerial vehicle (UAV) self-localization~\cite{Ye2021TIE}, target following~\cite{Li2016CVPRW}, and aerial cinematography~\cite{Bonatti2019IROS}. Driven by large-scale datasets~\cite{Real2017CVPR, Huang2021TPAMI, Fan2021IJCV} with the supervision of meticulous manual annotations, emerging deep trackers~\cite{Li2019CVPR, Cao_2021_ICCV, Chen2020CVPR, Guo2020CVPR} keep setting state-of-the-arts (SOTAs) in recent years. 

	\begin{figure}[!t]	
	\centering
	\subfloat[Qualitative comparison in typical night scenes.]
	{
		\includegraphics[width=0.97\linewidth]{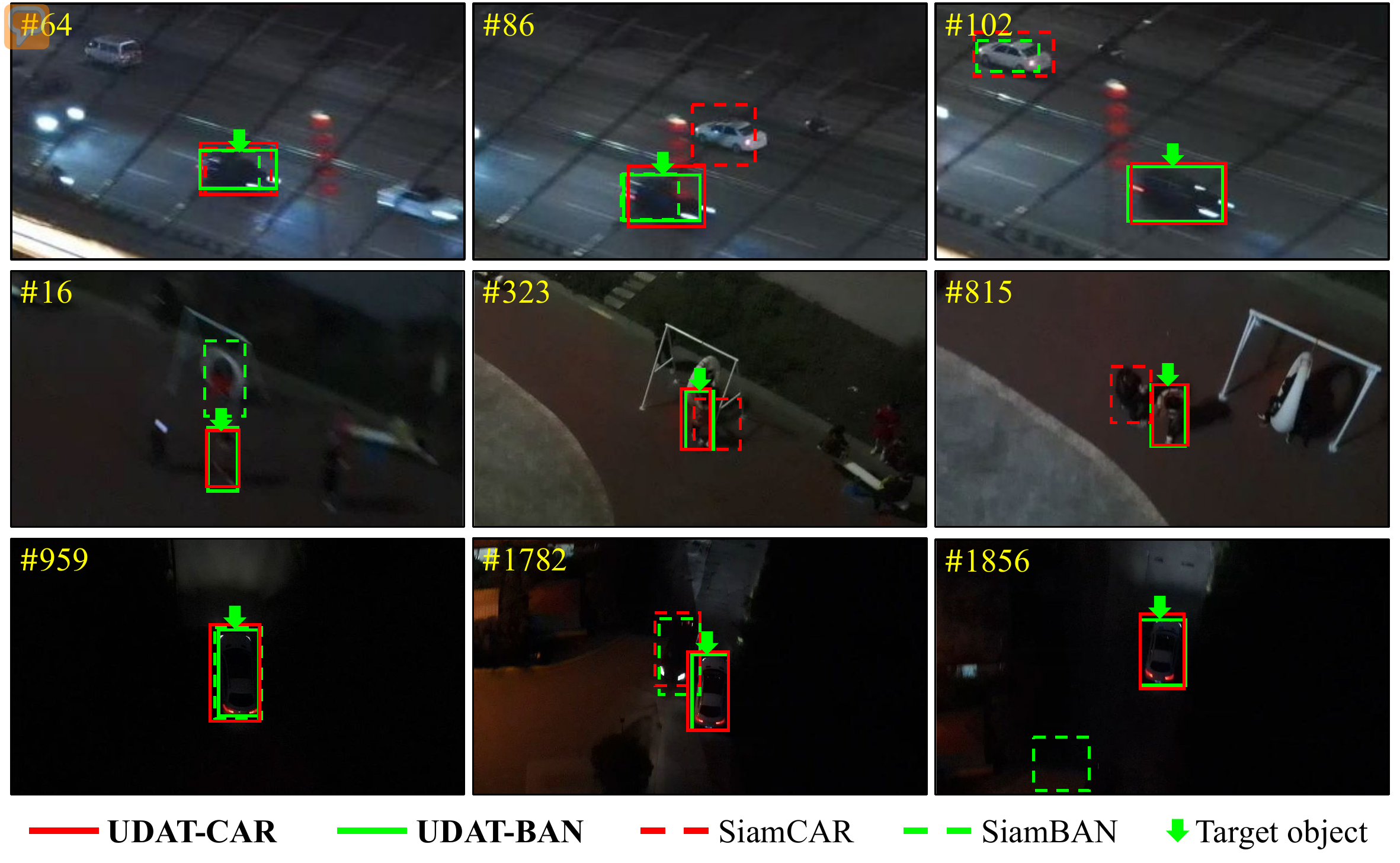}
	}
	
	\subfloat[Overall performance comparison on NAT2021-$test$.]
	{
		\includegraphics[width=0.97\linewidth]{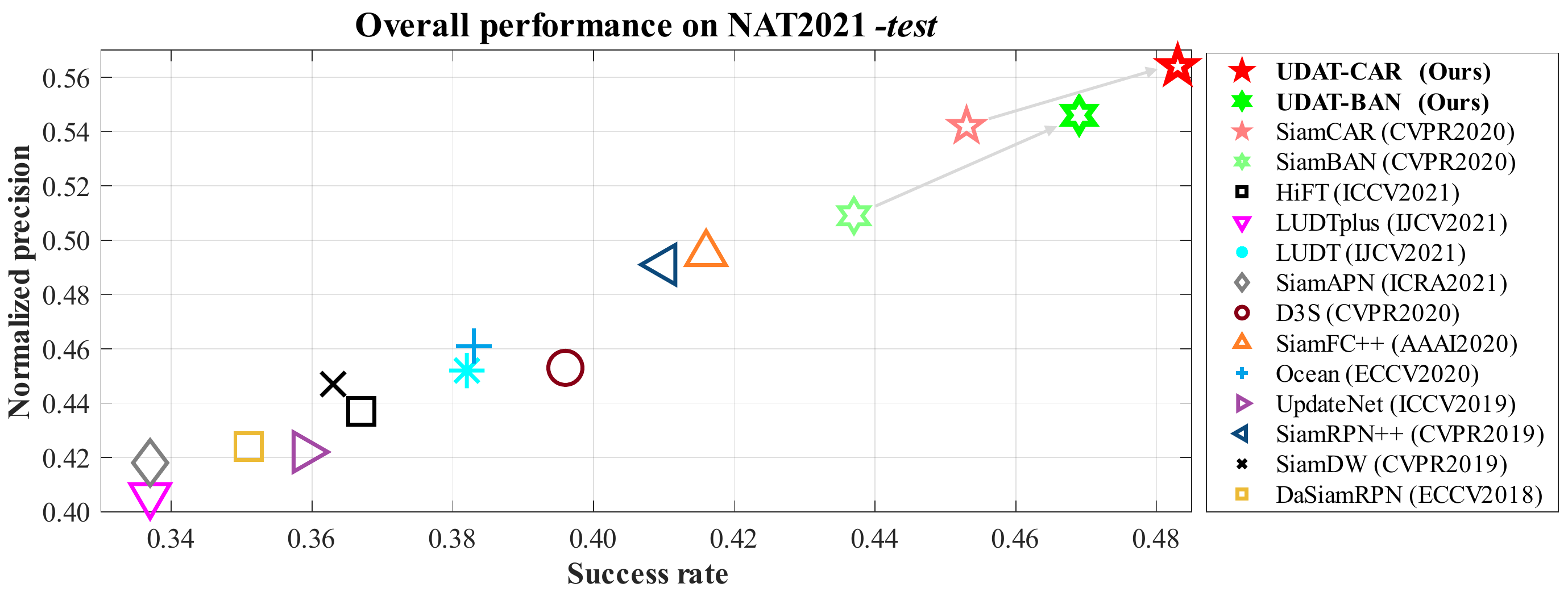}
	}
\setlength{\abovecaptionskip}{3pt}
	\caption
	{
	 (a) Qualitative comparison of the proposed unsupervised domain adaptive trackers (\ie, UDAT-CAR and UDAT-BAN) and their baselines~\cite{Guo2020CVPR, Chen2020CVPR}. (b) Overall performance of SOTA approaches on the constructed NAT2021-$test$ benchmark. The proposed UDAT effectively adapts general trackers to nighttime aerial tracking scenes and yields favorable performance.
	}
	\label{fig:fig1}
\vspace{-11pt}
\end{figure}	
	
	Despite the advances, whether current benchmarks or approaches are proposed for object tracking under favorable illumination conditions. 
	In contrast to daytime, images captured at night have low contrast, brightness, and signal-to-noise ratio (SNR).
	These differences cause the discrepancy in feature distribution of day/night images. 
	Due to the cross-domain discrepancy, current SOTA trackers generalize badly to nighttime scenes~\cite{Ye_2021_IROS, Ye_2022_RAL}, which severely impedes the broadening of relevant aerial robot applications.

	Regarding such domain gap and the performance drop, this work aims to address the cross-domain object tracking problem. In particular, we target adapting SOTA tracking models in daytime general conditions to nighttime aerial perspectives. One possible straightforward solution is to collect and annotate adequate target domain data for training. 
	Nevertheless, such a non-trivial workload is expensive and time-consuming, since backbones' pre-training alone generally takes millions of high-quality images~\cite{Deng2009CVPR}. We consequently consider the problem as an unsupervised domain adaptation task, where training data in the source domain is with well-annotated bounding boxes while that in the target domain has no manually annotated labels. Therefore, an \textbf{u}nsupervised \textbf{d}omain \textbf{a}daptive \textbf{t}racking framework, referred to as UDAT, is proposed for nighttime aerial tracking. To generate training patches of the target domain, we develop an object discovery strategy to explore potential objects in the unlabelled nighttime data. Besides, a bridging layer is proposed to bridge the gap of domain discrepancy for the extracted features.

	Furthermore, the feature domain is distinguished by virtue of a discriminator during adversarial learning.
	Drawing lessons from the huge potential of the Transformer~\cite{vaswani2017nips} in feature representation, both the bridging layer and the discriminator utilize a Transformer structure. 
	\Cref{fig:fig1} exhibits some qualitative comparisons of trackers adopting UDAT and the corresponding baselines. UDAT raises baselines' nighttime aerial tracking performance substantially. 
	Apart from methodology, we construct NAT2021, a benchmark comprising a \textit{test} set of 180 fully annotated video sequences and a \textit{train} set of over 276k unlabelled nighttime tracking frames, which serves as the first benchmark for unsupervised domain adaptive nighttime tracking. The main contributions of this work are fourfold:
	\begin{itemize}
		\item An unsupervised domain adaptive tracking framework, namely UDAT, is proposed for nighttime aerial tracking. To the best of our knowledge, the proposed UDAT is the first unsupervised adaptation framework for object tracking.
		\item A bridging layer and a day/night discriminator with Transformer structures are incorporated to align extracted features from different domains and narrow the domain gap between daytime and nighttime.
		\item A pioneering benchmark namely NAT2021, consisting of a fully annotated \textit{test} set and an unlabelled \textit{train} set, is constructed for domain adaptive nighttime tracking. An object discovery strategy is introduced for the unlabelled \textit{train} set preprocessing. 
		\item Extensive experiments on NAT2021-$test$ and the recent public UAVDark70~\cite{Li2021ICRA} benchmark verify the effectiveness and domain adaptability of the proposed UDAT in nighttime aerial tracking.
	\end{itemize}

	%-------------------------------------------------------------------------

	\begin{figure*}[!t]	
		\centering
		\includegraphics[width=0.85\linewidth]{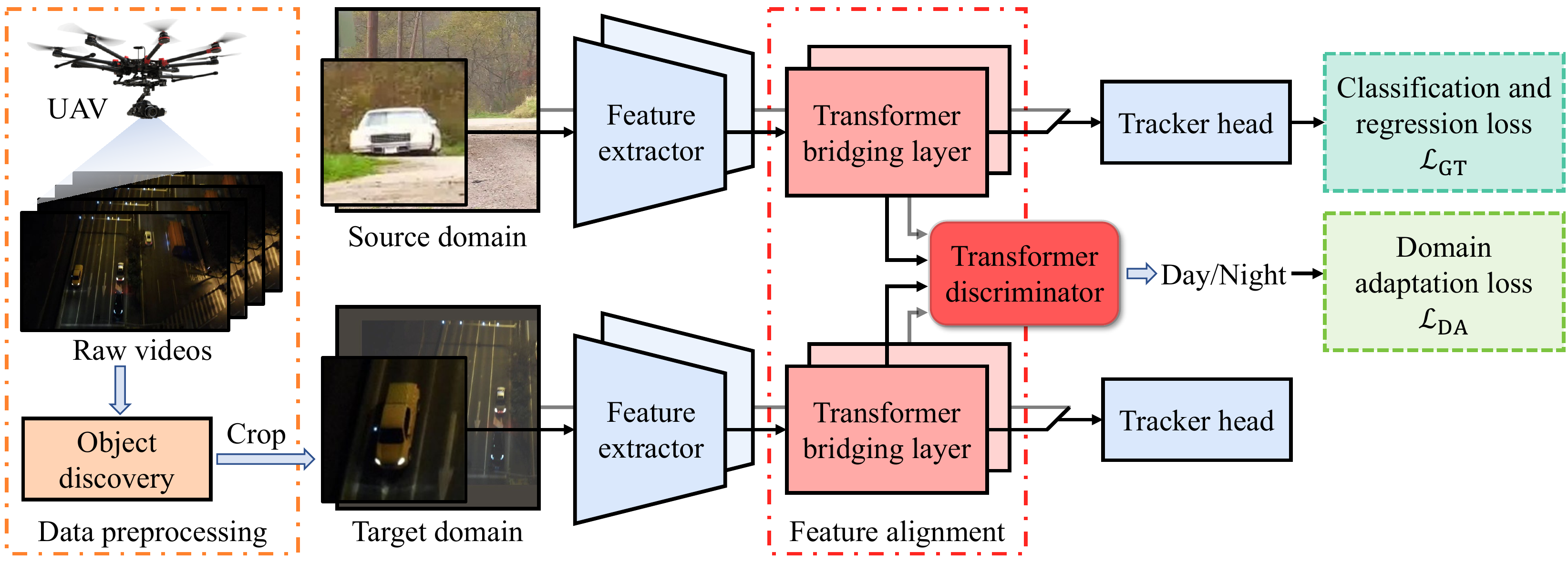}
			\setlength{\abovecaptionskip}{3pt}
		\caption
		{Illustration of the proposed unsupervised domain adaptation framework for nighttime aerial tracking. The object discovery module is employed to find potential objects in raw videos for training patch generation. Features extracted from different domains are aligned via the Transformer bridging layer. A Transformer day/night discriminator is trained to distinguish features between the source domain and the target domain.
		}  
		\label{fig:main}
		\vspace{-10pt}
	\end{figure*}
	
	\section{Related work}
	\subsection{Object tracking}
	Generally, recent object tracking approaches can be categorized as the discriminative correlation filter (DCF)-based approaches~\cite{Henriques2015TPAMI, Galoogahi2017ICCV, Huang2019ICCV, Li2020CVPR} and the Siamese network-based approaches~\cite{Li2019CVPR, Cao_2021_ICCV, Chen2020CVPR, Guo2020CVPR}. Due to the complicated online learning procedure, end-to-end training can be hardly realized on DCF-based trackers. Therefore, restricted to inferior handcrafted features or inappropriate deep feature extractors pre-trained for classification, DCF-based trackers lose their effectiveness in adverse conditions. 
	
	Benefiting from considerable training data and end-to-end learning, Siamese network-based trackers have achieved robust tracking performance. This line of approaches is pioneered by SINT~\cite{Tao2016CVPR} and SiamFC~\cite{Bertinetto2016ECCVW}, which regard object tracking as a similarity learning problem and train Siamese networks with large-scale image pairs. Drawing lessons from object detection, B. Li \etal~\cite{Li2018CVPR} introduce the region proposal network (RPN)~\cite{Ren2017TPAMI} into the Siamese framework. SiamRPN++~\cite{Li2019CVPR} further adopts a deeper backbone and feature aggregation architecture to improve tracking accuracy. To alleviate increasing hyperparameters along with the introduction of RPN, the anchor-free approaches~\cite{Chen2020CVPR, Guo2020CVPR, xu2020AAAI} adopt the per-pixel regression to directly predict four offsets on each pixel. Recently, Transformer~\cite{vaswani2017nips} is incorporated into the Siamese framework~\cite{Cao_2021_ICCV, Wang_2021_CVPR, Chen2021CVPR} to model global information and further boost tracking performance.

	Despite the great progress, object tracking in adverse conditions, for instance, nighttime aerial scenarios, lacks thorough study so far. In~\cite{Li2021ICRA}, a DCF framework integrated with a low-light enhancer is constructed while lacking transferability and being restricted to handcrafted features. Some studies~\cite{Ye_2021_IROS, Ye_2022_RAL} design tracking-related low-light enhancers for data preprocessing in the tracking pipeline. However, such a paradigm suffers from weak collaboration with the tracking model and the cascade structure can hardly learn to narrow the domain gap at the feature level.

	\subsection{Domain adaptation}
	Towards narrowing the domain discrepancy and transferring knowledge from the source domain to the target domain, domain adaptation attracts considerable attention and is widely adopted in image classification~\cite{sun2016aaai, Busto2017ICCV, Li2018TPAMI}. Beyond classification, Y. Chen \etal~\cite{Chen2018CVPR} design a domain adaptive object detection framework and tackle the domain shift on both image-level and instance-level. In~\cite{Huang_2018_ECCV}, an image transfer model is trained to perform day-to-night transformation for data augmentation before learning a detection model. Y. Sasagawa and H. Nagahara~\cite{Sasagawa2020ECCV} propose to merge a low-light image enhancement model and an object detection model to realize nighttime object detection. For the task of semantic segmentation, C. Sakaridis \etal~\cite{Sakaridis2020TPAMI} formulate a curriculum framework to adapt semantic segmentation models from day to night through an intermediate twilight domain. X. Wu \etal~\cite{Wu_2021_CVPR} employ an adversarial learning manner to train a domain adaptation network for nighttime semantic segmentation. S. Saha \etal~\cite{Saha_2021_CVPR} mine cross-task relationships and build a multi-task learning framework for semantic segmentation and depth estimation in the unsupervised domain adaptation setting. Despite the rapid development in other vision tasks, domain adaptation for object tracking has not been investigated yet. Therefore, an effective unsupervised domain adaptation framework for object tracking is urgently needed.
	
	%------------------------------------------------------------------------
	\section{Proposed method}
	\label{method}

	The paradigm of the proposed UDAT framework is illustrated in \cref{fig:main}. For data preprocessing of the unlabelled target domain, we employ a saliency detection-based strategy to locate potential objects and crop paired training patches. In the training pipeline, features generated by the feature extractor are modulated by the bridging layer. In this process, adversarial learning facilitates the reduction of feature distribution discrepancy between the source and target domains. Through this simple yet effective process, trackers can achieve pleasant efficiency and robustness for night scenes comparable to daytime tracking. 

	\begin{figure}[!t]	
		\centering
		\setlength{\abovecaptionskip}{5pt}
		\includegraphics[width=0.98\linewidth]{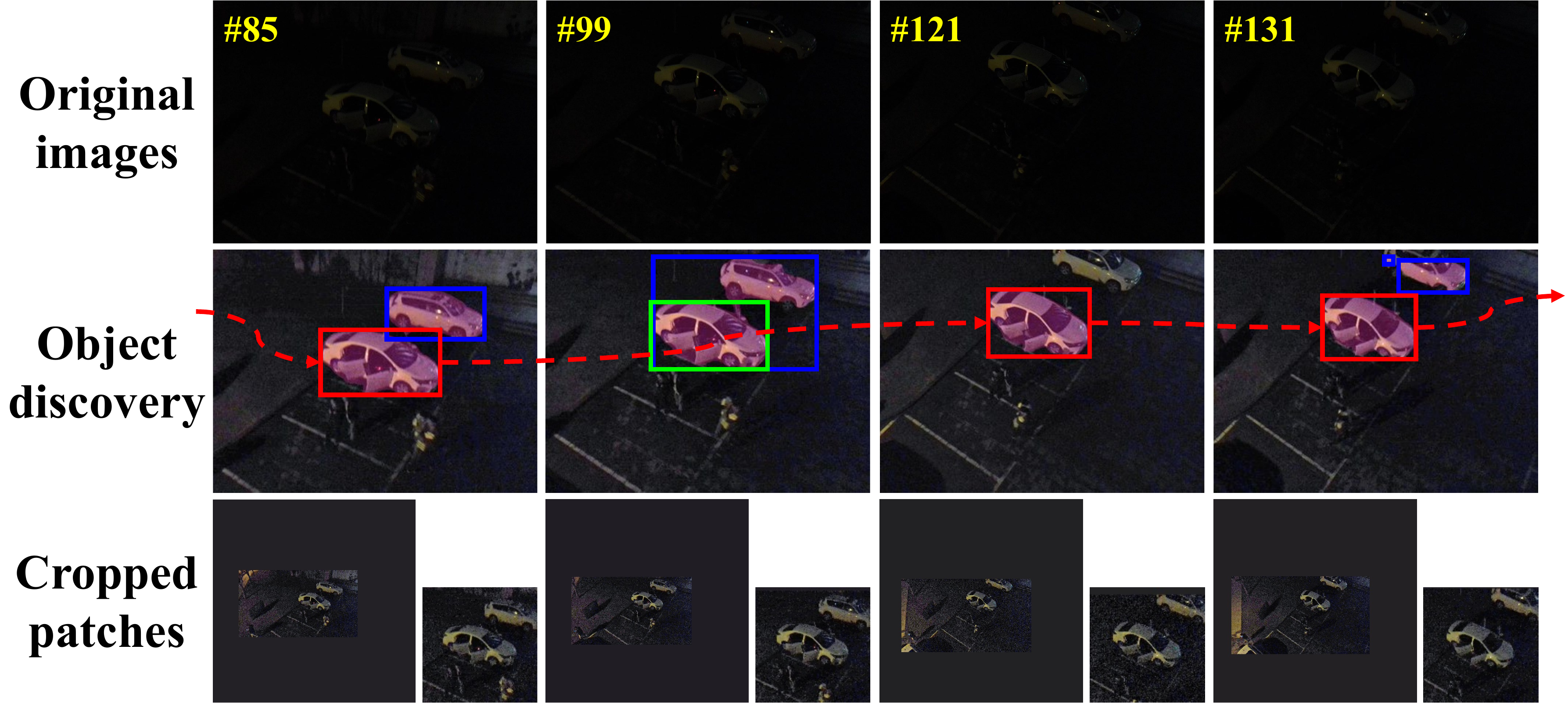}
		\caption
		{Illustration of object discovery, which contains low-light enhancement, salient object detection, and dynamic programming. The \pink{pink} masks indicate detected salient regions, while the boxes are circumscribed rectangles of these regions. Dynamic programming selects \NoOne{red} boxes and filters \NoThree{blue} ones. The \NoTwo{green} box is obtained by linear interpolation between two adjacent frames. Note that the cropped patches are enhanced for visualization, original patches are utilized in the practical training process instead.}
		\label{fig:data_precessing}
		\vspace{-12pt}
	\end{figure}
	
	\subsection{Data preprocessing}
	Since deep trackers take training patches as input in each training step, we develop an object discovery strategy for data preprocessing on the unlabelled train set. \Cref{fig:data_precessing} illustrates the preprocessing pipeline. The object discovery strategy involves three stages, \ie, low-light enhancement, salient object detection, and dynamic programming. Given the low visibility of nighttime images, original images are first lighted up by a low-light enhancer~\cite{Li2021TPAMI}. Afterward, enhanced images are fed into the video saliency detection model~\cite{Zhang_2021_ICCV}. Candidate boxes are then obtained by building the minimum bounding rectangle of detected salient regions. To generate a box sequence that locates the same object across the timeline, motivated by~\cite{Zheng2021ICCV}, we adopt dynamic programming to filter noisy boxes. Assuming two boxes from the $j$-th frame and the $k$-th frame as $[ {x}_{j, m},  {y}_{j, m},  {w}_{j, m},  {h}_{j, m}]$ and $[ {x}_{k, n},  {y}_{k, n},  {w}_{k, n},  {h}_{k, n}]$, where $m$ and $n$ indicate the box indexes, and $(x, y)$, $w$, $h$ denote the top-left coordinate, width, and height of the box, respectively, the normalized distance $D_{\rm norm}$ is obtained as:
	\begin{equation}
		\begin{split}
			D_{\rm norm} = &(\frac{ {x}_{j, m} -  {x}_{k, n}}{ {w}_{k, n}})^2 + (\frac{ {y}_{j, m} -  {y}_{k, n}}{ {h}_{k, n}})^2 \\&+ (\log(\frac{ {w}_{j, m}}{ {w}_{k, n}}))^2 + (\log(\frac{ {h}_{j, m}}{ {h}_{k, n}}))^2 \quad.
		\end{split}
	\end{equation}
	
	In dynamic programming, the normalized distance of candidate boxes between frames serves as a smooth reward, while adding a box in a frame to the box sequence means an incremental reward. For frames where none of the boxes is selected by dynamic programming, linear interpolation is adopted between the two closest frames to generate a new box. Ultimately, paired training patches are cropped from original images according to the obtained box sequence.
	
	\subsection{Network architecture}
	\noindent\textbf{Feature extractor.}
	Feature extraction of Siamese networks generally consists of two branches, \ie, the template branch and the search branch.
 	Both branches generate feature maps from the template patch $\mathbf{Z}$ and the search patch $\mathbf{X}$, namely $\varphi(\mathbf{Z})$ and $\varphi(\mathbf{X})$, by adopting an identical backbone network $\varphi$.  Generally, trackers adopt the last block or blocks of features for subsequent classification and regression, which can be represented as follows:
	\begin{equation}
		\begin{split}
			&\varphi(\mathbf{X})=\mathrm{Concat}(\mathcal{F}_{N-i}(\mathbf{X}),..., \mathcal{F}_{N-1}(\mathbf{X}), \mathcal{F}_{N}(\mathbf{X})) \ ,\\
			&\varphi(\mathbf{Z})=\mathrm{Concat}(\mathcal{F}_{N-i}(\mathbf{Z}),..., \mathcal{F}_{N-1}(\mathbf{Z}), \mathcal{F}_{N}(\mathbf{Z})) \ ,
		\end{split}
	\end{equation}
	where $\mathcal{F}_{*}(\cdot)$ indicates features extracted from the $*$-th block of a backbone with $N$ blocks in total, and $\mathrm{Concat}$ denotes channel-wise concatenation. Since both $\varphi(\mathbf{X})$ and $\varphi(\mathbf{Z})$ will pass through the following Transformer bridging layer and discriminator, we take the instance of $\varphi(\mathbf{X})$ in the following introduction for clarity.

	\noindent\textbf{Transformer bridging layer.}
	Features extracted from daytime and nighttime images are with a huge gap, such domain discrepancy leads to inferior tracking performance at night. Before feeding the obtained features to the tracker head for object localization, we propose to bridge the gap between the feature distributions through a bridging layer. In consideration of the strong modeling capability of the Transformer~\cite{vaswani2017nips} for long-range inter-independencies, we design the bridging layer with a Transformer structure. Taking the search branch as instance, positional encodings $\mathbf{P}$ are first added to the input feature $\varphi(\mathbf{X}) \in \mathbb{R}^{N \times H \times W}$. Next, the summation is flattened to $(\mathbf{P}+\varphi(\mathbf{X})) \in \mathbb{R}^{HW \times N}$. Multi-head self-attention (MSA) is then conducted as: 
	\begin{equation}
		\begin{aligned}
			&\widehat{ {\varphi(\mathbf{X})}}'=\mathrm{MSA}( {\mathrm{\mathbf{P}+\varphi(\mathbf{X})}})+ {\mathrm{\mathbf{P}+\varphi(\mathbf{X})}}\quad,\\
			&\widehat{ {\varphi(\mathbf{X})}}=\mathrm{LN}(\mathrm{FFN}(\mathrm{Mod}(\mathrm{LN}( \widehat{ {\varphi(\mathbf{X})}}')))+ \widehat{ {\varphi(\mathbf{X})}}')\quad,
		\end{aligned}
	\end{equation}
	where $\widehat{ {\varphi(\mathbf{X})}}'$ is an intermediate variable and $\mathrm{LN}$ indicates layer normalization. Moreover, $\mathrm{FFN}$ denotes the fully connected feed-forward network, which consists of two linear layers with a ReLU in between. $\mathrm{Mod}$ is a modulation layer in~\cite{Cao_2021_ICCV} to fully explore internal spatial information. The final output is flattened back to $N\times H \times W$. For each head of MSA, the attention function can be formulated as:
	\begin{equation}
		\mathrm{Attention}(\mathbf{Q},\mathbf{K},\mathbf{V})=\mathrm{Softmax}(\frac{\mathbf{QK}^T}{\sqrt{d_k}})\mathbf{V} \quad.
	\end{equation}
	
	In our case, the queries $\mathbf{Q}$, keys $\mathbf{K}$, and values $\mathbf{V}$ are equal to the product of $(\mathbf{P}+\varphi(\mathbf{X}))$ and the corresponding projection matrix. 
	By virtue of superior information integration of self-attention, the proposed Transformer bridging layer is adequate to modulate the nighttime object features output by the backbone for effective similarity maps. 

\begin{figure}[!t]	
		\centering
		\includegraphics[width=0.98\linewidth]{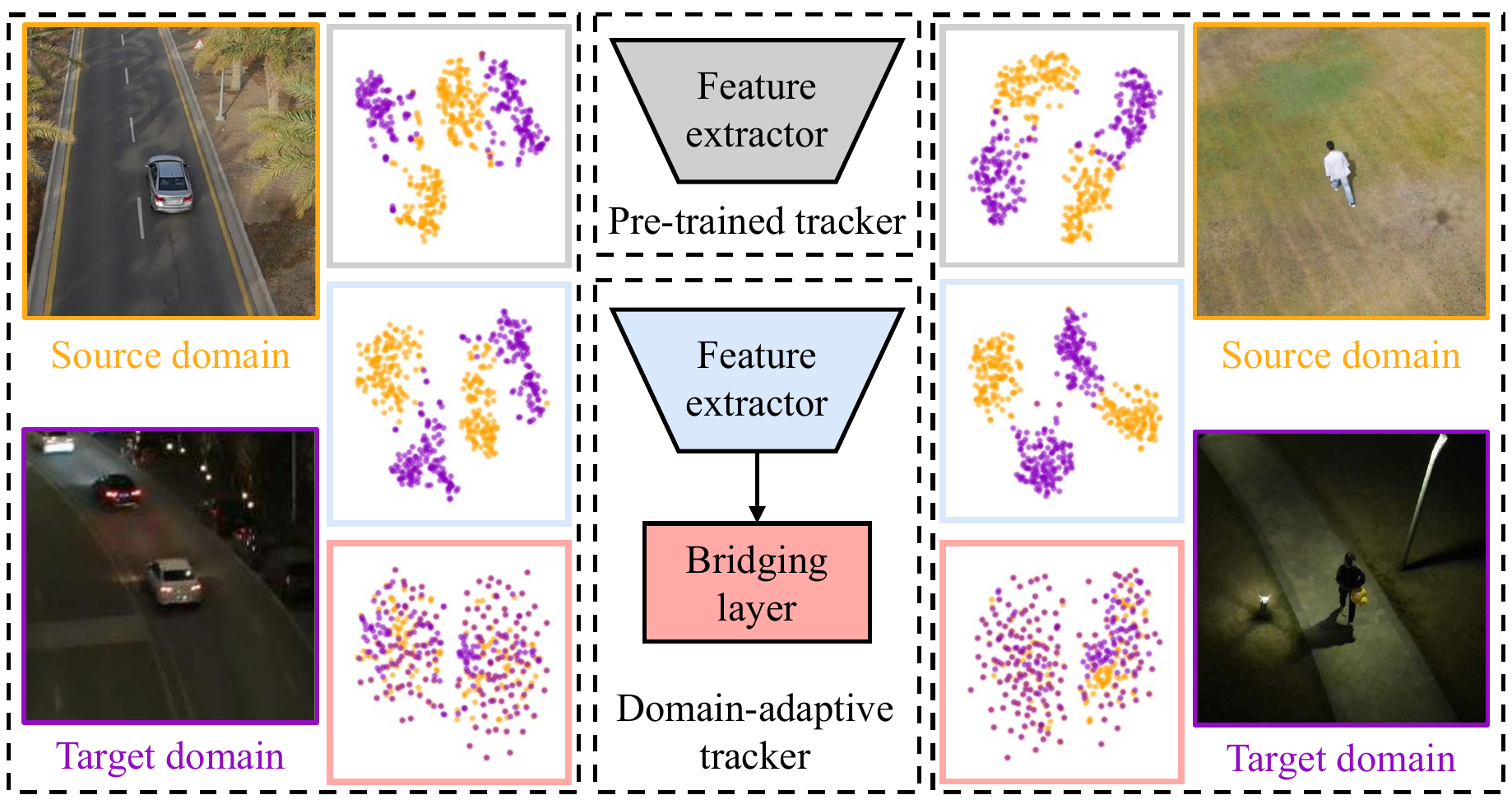}
			\setlength{\abovecaptionskip}{3pt}
		\caption
		{Feature visualization by t-SNE~\cite{van2008visualizing} of day/night similar scenes. \NoFour{Gold} and \purple{purple} indicate source and target domains, respectively. The scattergrams from top to down depict day/night features from feature extractor in the pre-trained tracker, feature extractor in the domain-adaptive tracker, and the bridging layer. The proposed Transformer bridging layer effectively narrows domain discrepancy.
		}  
		\label{fig: feature}
		\vspace{-8pt}
	\end{figure}

	\Remark \Cref{fig: feature} displays the t-SNE~\cite{van2008visualizing} visualizations of features from feature extractor in the baseline, feature extractor in the domain-adaptive tracker, and the bridging layer. From which we can observe that features extracted by backbones have a clear discrepancy, while those modified by the bridging layer show a coincidence in distribution.
	
	\noindent\textbf{Transformer discriminator.}
	The proposed UDAT framework is trained in an adversarial learning manner. A day/night feature discriminator $\mathrm{D}$ is designed to facilitate aligning the source and target domain features, which consists of a gradient reverse layer (GRL)~\cite{Ganin2015PMLR} and two Transformer layers. Given the modulated feature map $\widehat{ {\varphi(\mathbf{X})}}$, the softmax function is performed and followed by a GRL:
	\begin{equation}
		\mathbf{F} = \mathrm{GRL}(\mathrm{Softmax}(\widehat{ {\varphi(\mathbf{X})}})) \quad.
	\end{equation}
	
	The intermediate feature $\mathbf{F} \in \mathbb{R}^{N \times H \times W}$ is then passed through a convolution layer with a kernel size of $4\times 4$ and stride of 4 for patch embedding. $\mathbf{F}$ is then flattened to $(\frac{H}{4} \times \frac{W}{4}) \times N$ and concatenated with a classification token~$\mathbf{c}$ as:
	\begin{equation}
		\mathbf{F'} = \mathrm{Concat}(\mathbf{c}, \mathrm{Flat}(\mathrm{Conv}(\mathbf{F}))) \quad.
	\end{equation}
	
	Afterward, $\mathbf{F'}$ is input to two Transformer layers. Ultimately, the classification token $\mathbf{c}$ is regarded as the final predicted results.
	In the adversarial learning process, the discriminator is optimized to distinguish whether the features are from the source domain or the target domain correctly. 
	
	\noindent\textbf{Tracker head.} After the Transformer bridging layer, cross-correlation operation is conducted on the modulated features $\widehat{\varphi(\mathbf{X})}$ and $\widehat{\varphi(\mathbf{Z})}$ to generate a similarity map. Finally, the tracker head performs the classification and regression process to predict the object position.

	\subsection{Objective functions}
	
	\noindent\textbf{Classification and regression loss.}
	In the source domain training line, the classification and regression loss $\mathcal{L}_{\mathrm{GT}}$ between the ground truth and the predicted results are applied to ensure the normal tracking ability of trackers. We adopt tracking loss consistent with the baseline trackers without modification.
	
	\noindent\textbf{Domain adaptation loss.}
	In adversarial learning, the least-square loss function~\cite{Mao_2017_ICCV} is introduced to train the generator~$G$, aiming at generating source domain-like features from target domain images to fool the discriminator $\mathrm{D}$ while frozen. Here the generator $G$ can be regarded as the feature extractor along with the Transformer bridging layer. Considering both the template and search features, the adversarial loss is described as follows:
	\begin{equation}
		\mathcal{L}_{\mathrm{adv}} = (\mathrm{D}(\widehat{ {\varphi(\mathbf{X}_{\rm t})}})-\ell_{\rm s})^2+(\mathrm{D}(\widehat{ {\varphi(\mathbf{Z}_{\rm t})}}))-\ell_{\rm s})^2
		\quad,
	\end{equation}
	where $\rm s$ and $\rm t$ refer to the source and the target domains, respectively. Besides, $\ell_{\rm s}$ denotes the label for the source domain, which has the same size as the output of $\mathrm{D}$. In summary, the total training loss for the tracking network is defined as:
	\begin{equation}
		\mathcal{L}_{\rm total} = \mathcal{L}_{\rm GT}+\lambda\mathcal{L}_{\rm adv}
		\quad,
	\end{equation}
	where $\lambda$ is a weight to balance the loss terms. We set $\lambda$ as 0.01 in implementation.
	
	During the training process, the tracking network and discriminator $\mathrm{D}$ are optimized alternatively. We define the loss function of $\mathrm{D}$ as:
	 	\begin{equation}
	 	L_{\rm D} = \sum_{d= \rm s,t}(\mathrm{D}(\widehat{ {\varphi(\mathbf{X}_{d})}})-\ell_d)^2+(\mathrm{D}(\widehat{ {\varphi(\mathbf{Z}_{d})}})-\ell_d)^2 \quad.
	 	\end{equation}
	
	Trained with true domain labels of input features, $\mathrm{D}$ learns to discriminate feature domains efficiently.
	%------------------------------------------------------------------------

	\section{NAT2021 benchmark}
	\label{sec:data}
	The nighttime aerial tracking benchmark, namely NAT2021, is developed to give a comprehensive performance evaluation of nighttime aerial tracking and provide adequate unlabelled nighttime tracking videos for unsupervised training. Compared to the existing nighttime tracking benchmark~\cite{Li2021ICRA} in literature, NAT2021 stands a two times larger \textit{test} set, an unlabelled \textit{train} set, and novel illumination-oriented attributes.
	\subsection{Sequence collection}
\begin{figure}[!t]	
	\centering
	\setlength{\abovecaptionskip}{5pt}
	\includegraphics[width=0.99\linewidth]{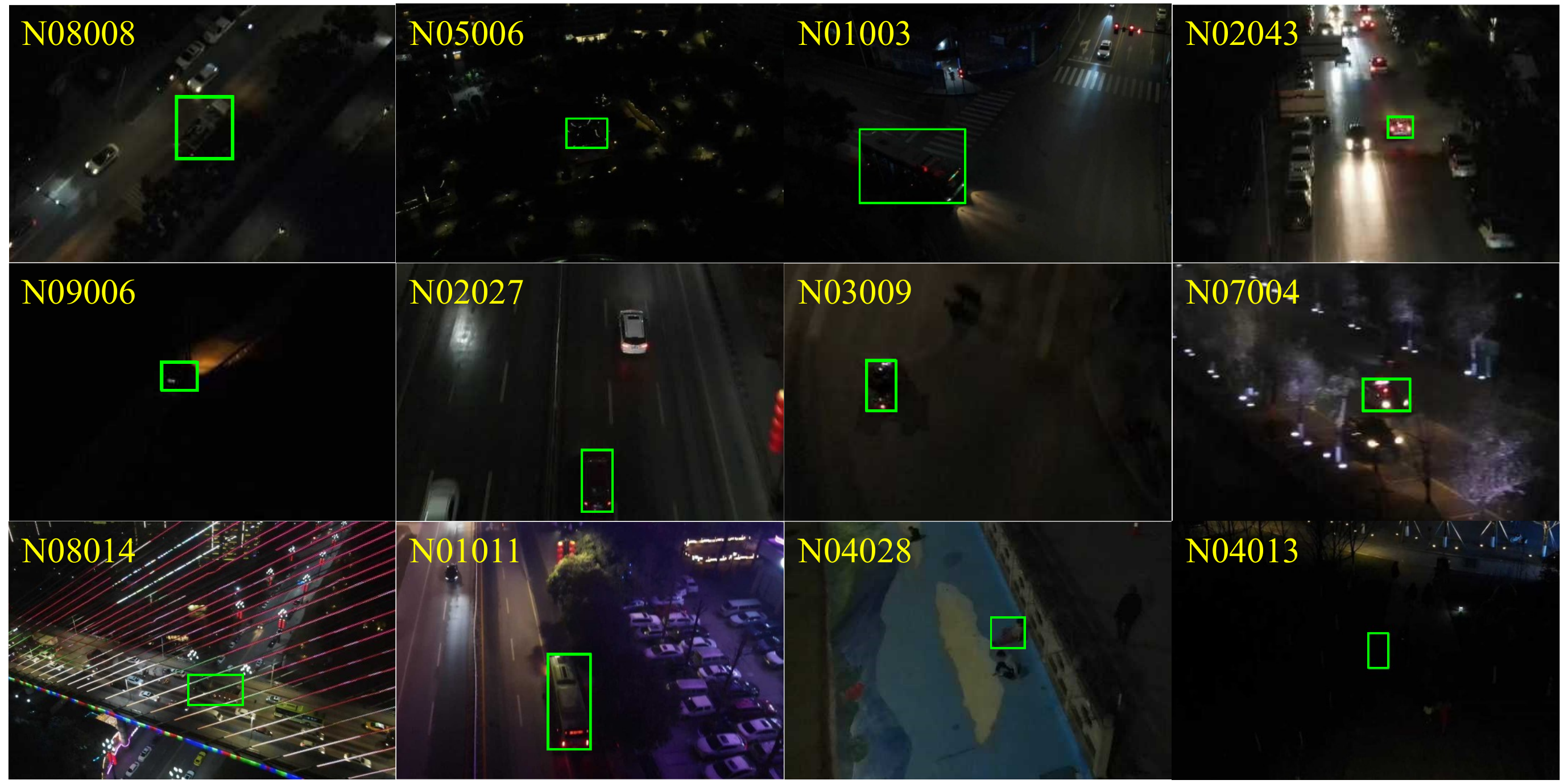}
	\caption
	{First frames of selected sequences from NAT2021-$test$. The green boxes mark the tracking objects, while the top-left corner of the images display sequence names.}  
	\label{fig:object}
	\vspace{-5pt}
\end{figure}

	Images in NAT2021 are captured in diverse nighttime scenes (\eg, roads, urban landscapes, and campus) by a DJI Mavic Air 2 UAV\footnote{More information of the UAV can be found at \url{https://www.dji.com/cn/mavic-air-2}.} in a frame rate of 30 frames/s. Sequence categories consist of a wide variety of targets (\eg, cars, trucks, persons, groups, buses, buildings, and motorcycles) or activities (\eg, cycling, skating, running, and ball games). Consequently, the \textit{test} set contains 180 nighttime aerial tracking sequences with more than 140k frames in total, namely NAT2021-$test$. \Cref{fig:object} displays some first frames of selected sequences. In order to provide an evaluation of long-term tracking performance, we further build a long-term tracking subset namely NAT2021-$L\text{-}test$ consisting of 23 sequences that are longer than 1400 frames. Moreover, the training set involves 1400 unlabelled sequences with over 276k frames totally, which is adequate for the domain adaptive tracking task. A statistical summary of NAT2021 is presented in \cref{tab:benchmark}. 
	
	\Remark Sequences in NAT2021 are recorded by ourselves using UAVs with permission. No personally identifiable information or offensive content is involved.
	\subsection{Annotation}
	The frames in NAT2021-$test$ and NAT2021-$L\text{-}test$ are all manually annotated by annotators familiar with object tracking. For accuracy, the annotation process is conducted on images enhanced by a low-light enhancement approach~\cite{Li2021TPAMI}.  Afterward, visual inspection by experts and bounding box refinement by annotators are conducted iteratively for several rounds to ensure high-quality annotation.

\begin{table}[!b]
	\vspace{-8pt}
	\renewcommand\tabcolsep{3pt}
	\centering
	\caption{Statistics of NAT2021. $test$: test set; $L\text{-}test$: long-term tracking test set; $train$: train set.}
	\scriptsize
	\begin{tabular}{lccc}
		\toprule
		& NAT2021-$test$ & NAT2021-$L\text{-}test$ & NAT2021-$train$ \\
		\midrule
		Videos & 180    & 23     & 1,400 \\
		Total frames & 140,815 & 53,564 & 276,081 \\
		Min frames & 81     & 1,425  & 30 \\
		Max frames & 1,795  & 3,866  & 345 \\
		Avg. frames & 782    & 2,329  & 197 \\
		Manual annotation & \checkmark & \checkmark &  \\
		\bottomrule
	\end{tabular}%
%\vspace{-10pt}
	\label{tab:benchmark}%
\end{table}%	

	\begin{figure}[!t]	
	\centering
	\setlength{\abovecaptionskip}{3pt}
	\includegraphics[width=0.98\linewidth]{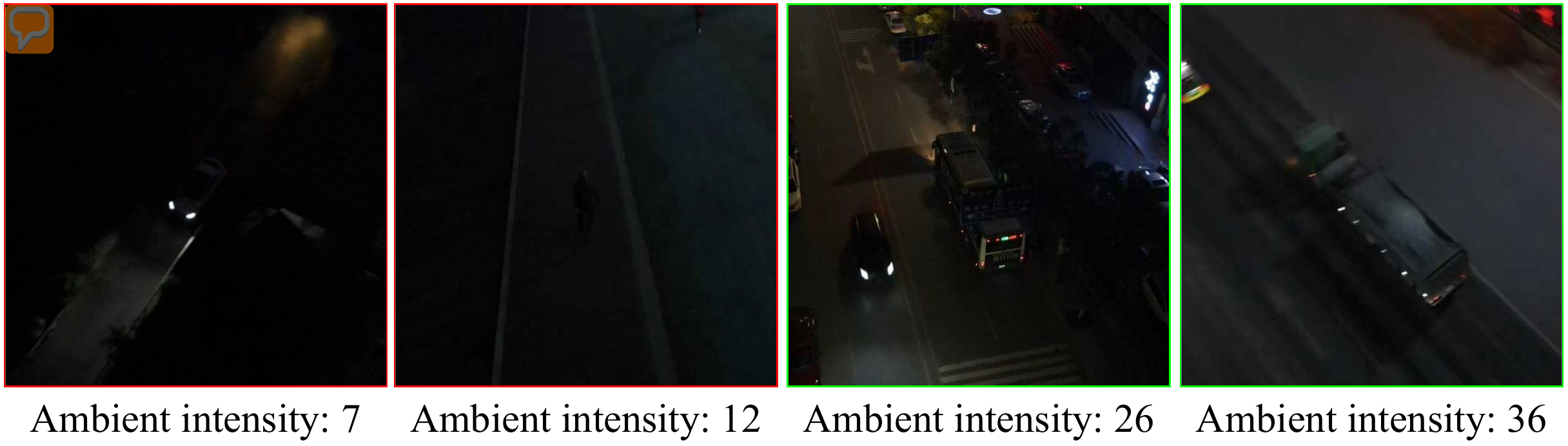}
	\caption
	{Ambient intensity of some scenarios. With an average ambient intensity of less than 20, objects are hard to distinguish with naked eyes. Such sequences are annotated with the low ambient intensity attribute.}  
	\label{fig:intensity}
%	\vspace{-10pt}
\end{figure}	

\begin{figure*}[!t]	
	\centering
	\subfloat[Precision, normalized precision, and success plots on NAT2021-$test$.]
	{
		\includegraphics[width=0.325\linewidth]{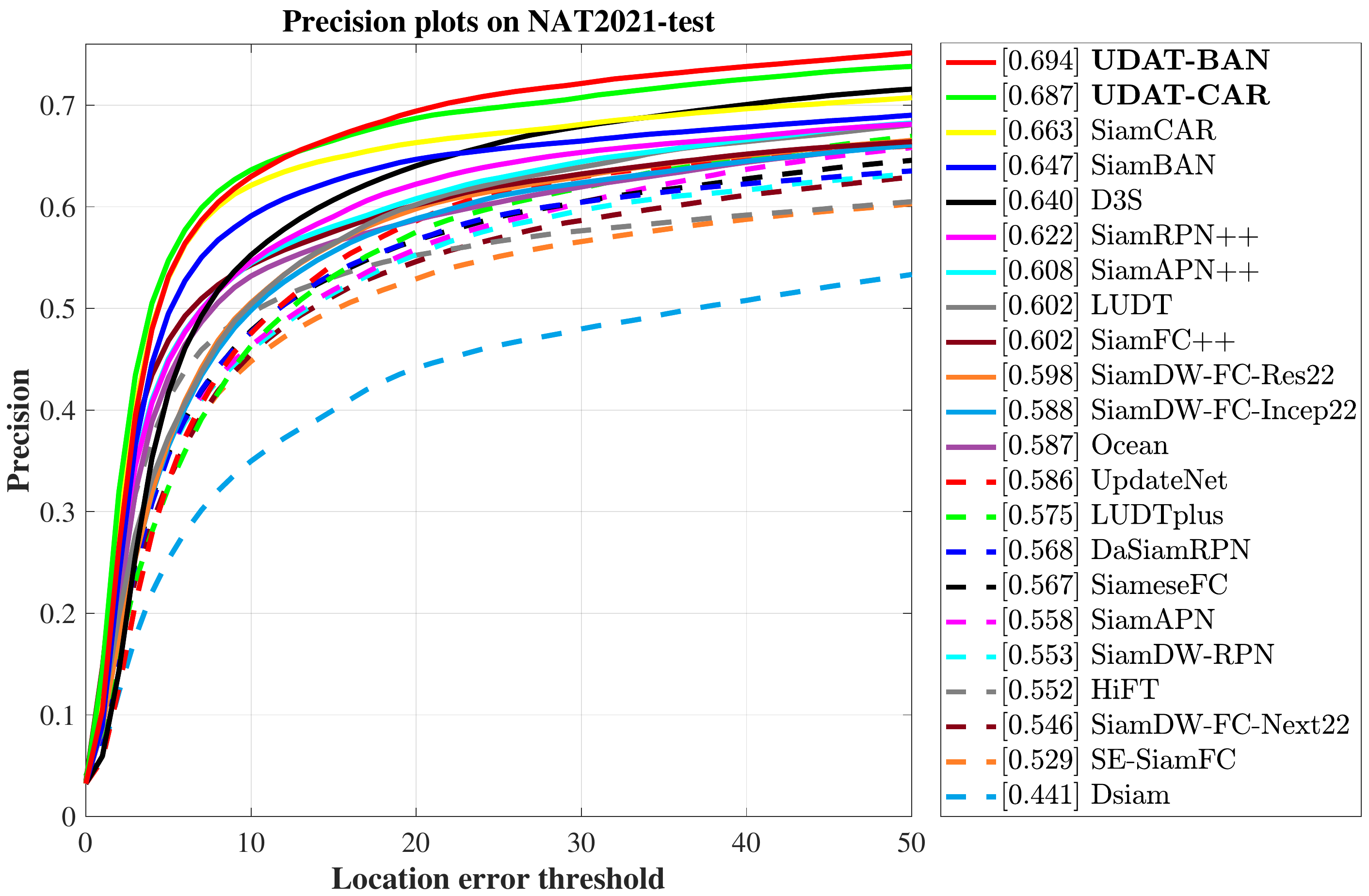}
		\includegraphics[width=0.325\linewidth]{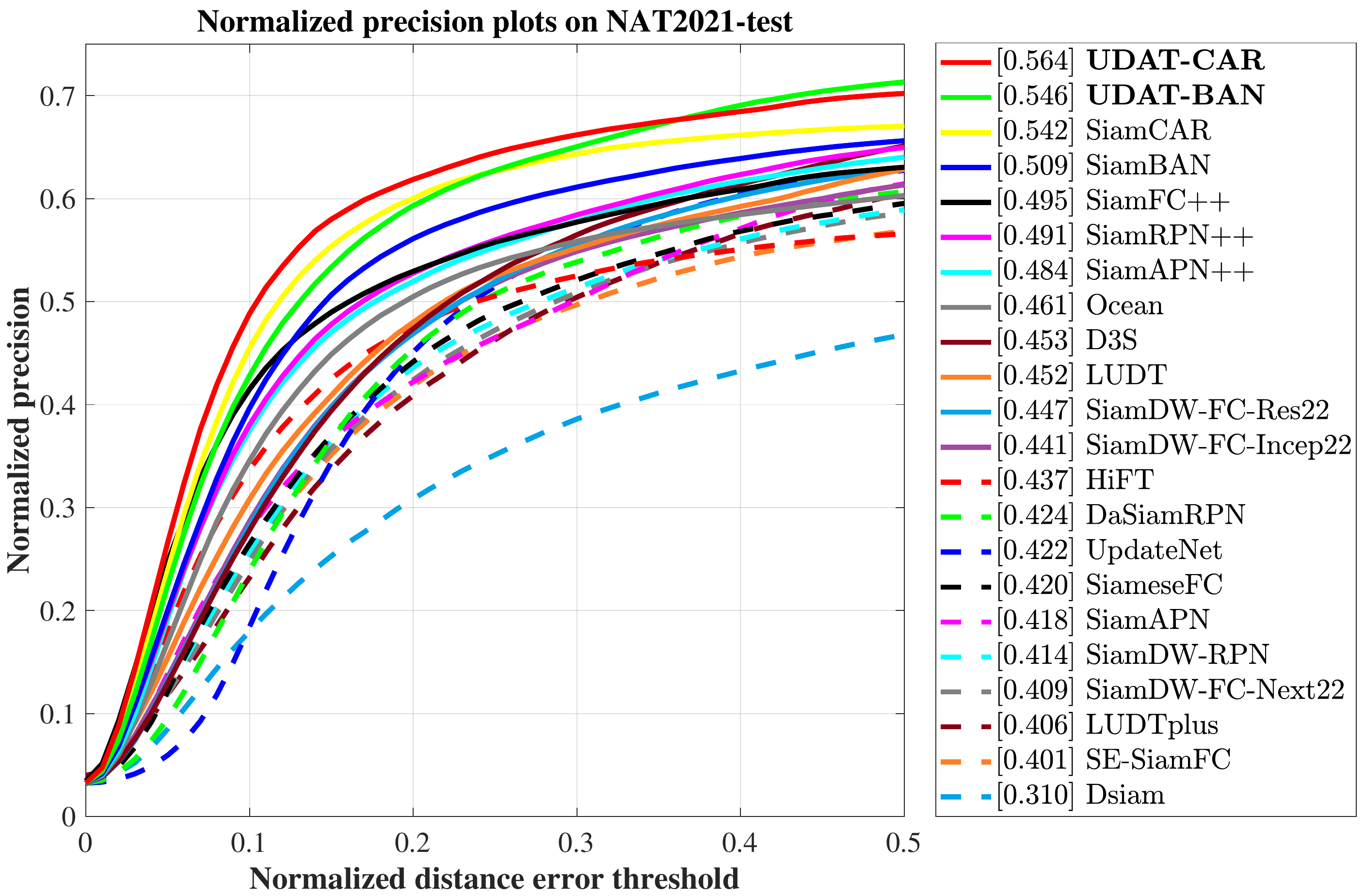}
		\includegraphics[width=0.325\linewidth]{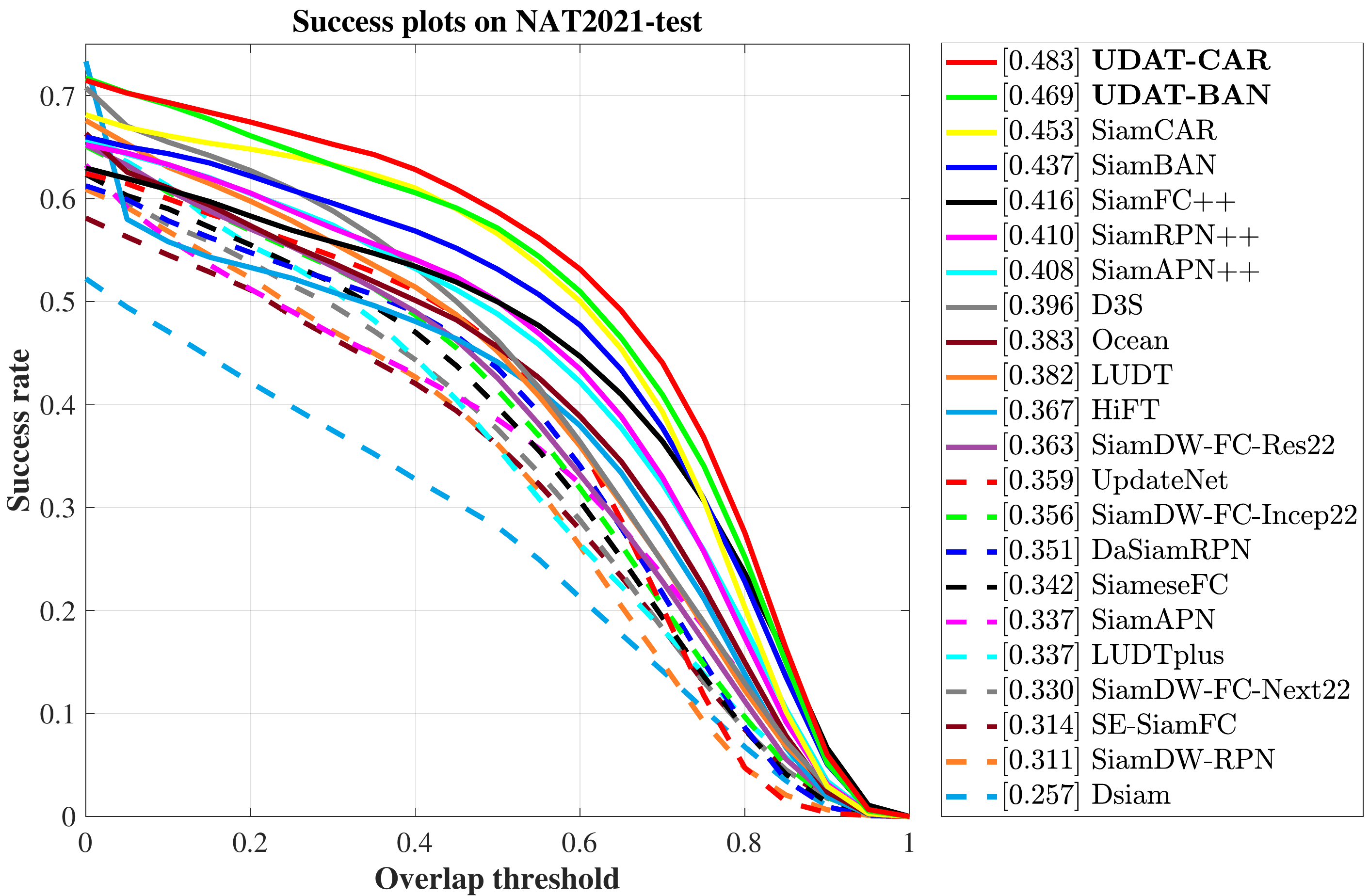}
	}
	
	\subfloat[Precision, normalized precision, and success plots on UAVDark70.]
	{
		\includegraphics[width=0.325\linewidth]{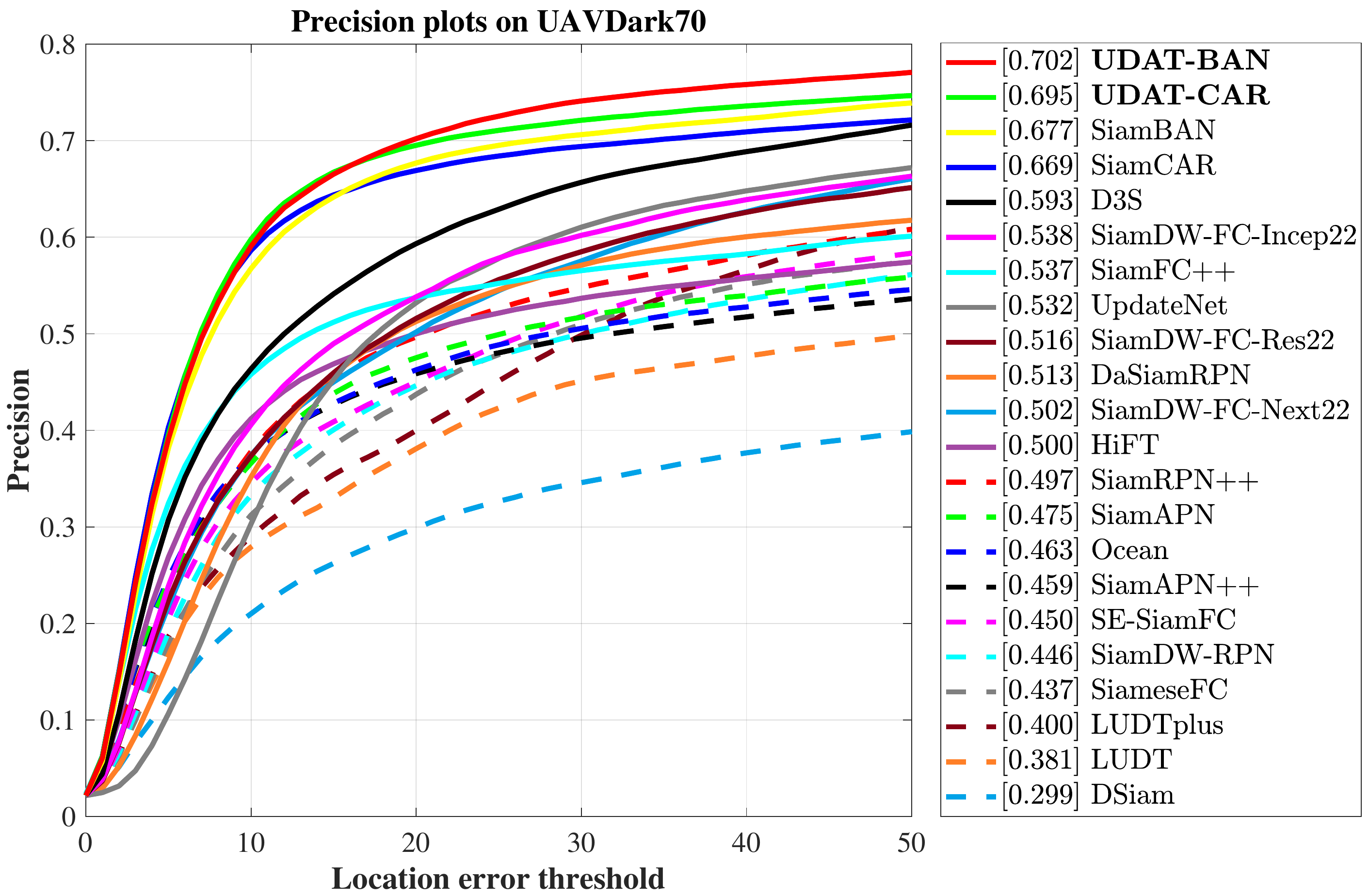}
		\includegraphics[width=0.325\linewidth]{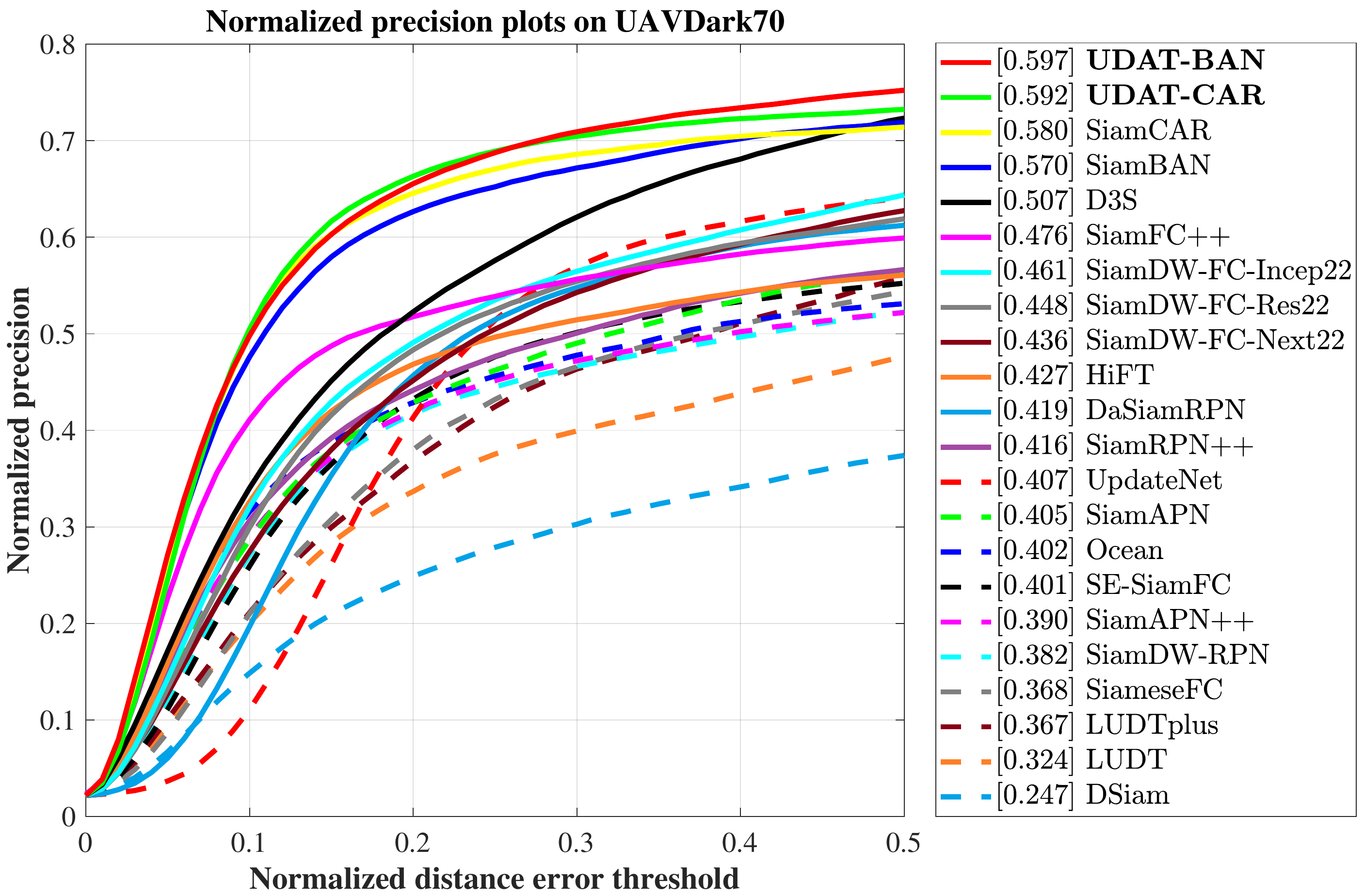}
		\includegraphics[width=0.325\linewidth]{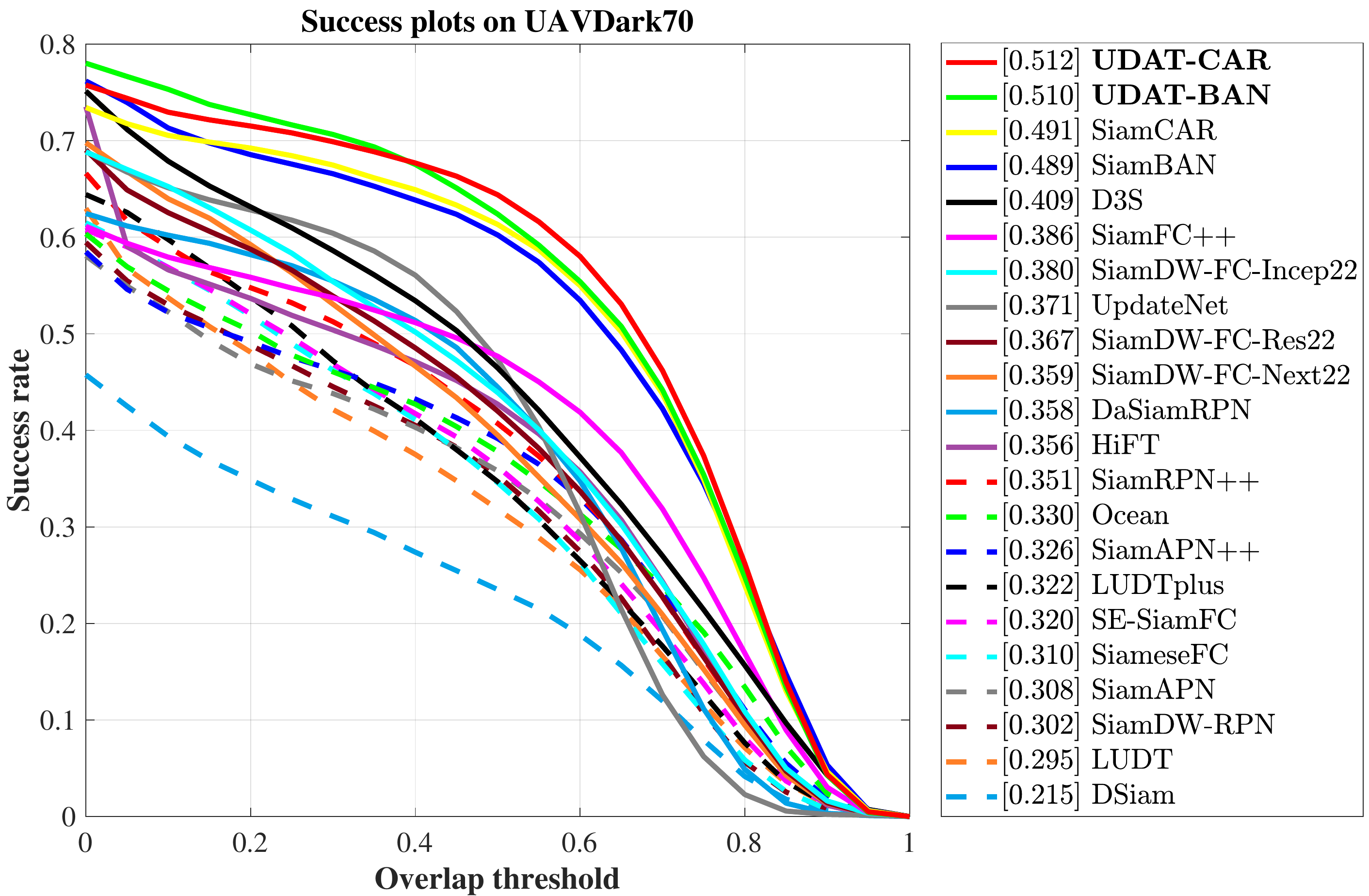}
	}
	\setlength{\abovecaptionskip}{3pt}
	\caption
	{
		Overall performance of SOTA trackers and UDAT on nighttime aerial tracking benchmarks. The results show that the proposed UDAT trackers realize top-ranked performance and improve baseline trackers favorably.
	}
	\label{fig:all}
	\vspace{-5pt}
\end{figure*}		

	\subsection{Attributes}
	To give an in-depth analysis of trackers, we annotate the test sequences into 12 different attributes, including aspect ratio change (ARC), background clutter (BC), camera motion (CM), fast motion (FM), partial occlusion (OCC), full occlusion (FOC), out-of-view (OV), scale variation (SV), similar object (SOB), viewpoint change (VC), illumination variation (IV), and low ambient intensity (LAI). In particular, to take a closer look at how illumination influences trackers, we rethink and redesign the illumination-related attributes. Concretely, the average pixel intensity of the local region centered at the object is computed and regarded as the illuminance intensity of the current frame. The average illuminance level of a sequence is considered the ambient intensity of the tracking scene. Sequences with different ambient intensities are displayed in \cref{fig:intensity}, we observe that objects are hard to distinguish with naked eyes with an ambient intensity of less than 20. Therefore, such sequences are labelled with the LAI attribute. 
	
	\Remark In contrast to annotating the attribute of IV intuitively as previous tracking benchmarks do, this work judges IV according to the maximum difference of the illuminance intensity across a tracking sequence. More details of the attributes can be found in \textit{supplementary material}.
	
	Moreover, we evaluate current top-ranked trackers on the proposed benchmark (see \cref{sec: eval}), the results show that SOTA trackers can hardly yield satisfactory performance at a nighttime aerial view as in daytime benchmarks.

	\section{Experiments}
	\subsection{Implementation details}
	We implement our UDAT framework using PyTorch on an NVIDIA RTX A6000 GPU. The discriminator is trained using the Adam~\cite{Kingma2015ICLR} optimizer. The base learning rate is set to 0.005 and is decayed following the poly learning rate policy with a power of 0.8. The bridging layer adopts a base learning rate of 0.005 and is optimized with the baseline tracker. The whole training process lasts 20 epochs. The top-ranked trackers~\cite{Guo2020CVPR, Chen2020CVPR} in the proposed benchmark are adopted as baselines. To achieve faster convergence, tracking models pre-trained on general datasets~\cite{russakovsky2015imagenet, Real2017CVPR, lin2014coco, Huang2021TPAMI, Fan2021IJCV} are served as the baseline models. For fairness, tracking datasets~\cite{Huang2021TPAMI, Real2017CVPR} that the pre-trained models learned on are adopted and no new datasets are introduced in the source domain. We adopt the one-pass evaluation (OPE) and rank performances using success rate, precision, and normalized precision. Evaluation metric definitions and more experiments can be found in the \textit{supplementary material}.
	
\begin{table}[!b]
	\setlength{\belowcaptionskip}{-10pt} 
	\renewcommand\tabcolsep{4pt}
	\centering
	\scriptsize
	\caption{Performance comparison of UDAT and baseline trackers. $\Delta$ denotes gains of percentages brought by UDAT.}
%	\vspace{-5pt}
	\begin{tabular}{lcccccc}
		\toprule
		\multicolumn{1}{c}{\multirow{2}[2]{*}{Benchmarks}} & \multicolumn{3}{c}{NAT2021-$test$} & \multicolumn{3}{c}{UAVDark70} \\
		& Prec.  & Norm. Prec.     & Succ.  & Prec.  & Norm. Prec.     & Succ. \\
		\midrule
		SiamCAR & 0.663  & 0.542  & 0.453  & 0.669  & 0.580  & 0.491 \\
		UDAT-CAR & 0.687  & 0.564  & 0.483  & 0.695  & 0.592  & 0.512 \\
		$\Delta_{\rm CAR}$  (\%) & \textbf{3.62} & \textbf{4.06} & \textbf{6.62} & \textbf{3.89} & \textbf{2.07} & \textbf{4.28} \\
		\midrule
		SiamBAN & 0.647  & 0.509  & 0.437  & 0.677  & 0.570  & 0.489 \\
		UDAT-BAN & 0.694  & 0.546  & 0.469  & 0.702  & 0.597  & 0.510 \\
		$\Delta_{\rm BAN}$  (\%) & \textbf{7.26} & \textbf{7.27} & \textbf{7.32} & \textbf{3.69} & \textbf{4.74} & \textbf{4.29} \\
		\bottomrule
	\end{tabular}%
	\label{tab:pro}%
\end{table}%

	\begin{figure*}[!t]	
	\centering
	\subfloat[Illumination variation on NAT2021-$test$.]
	{
		\includegraphics[width=0.242\linewidth]{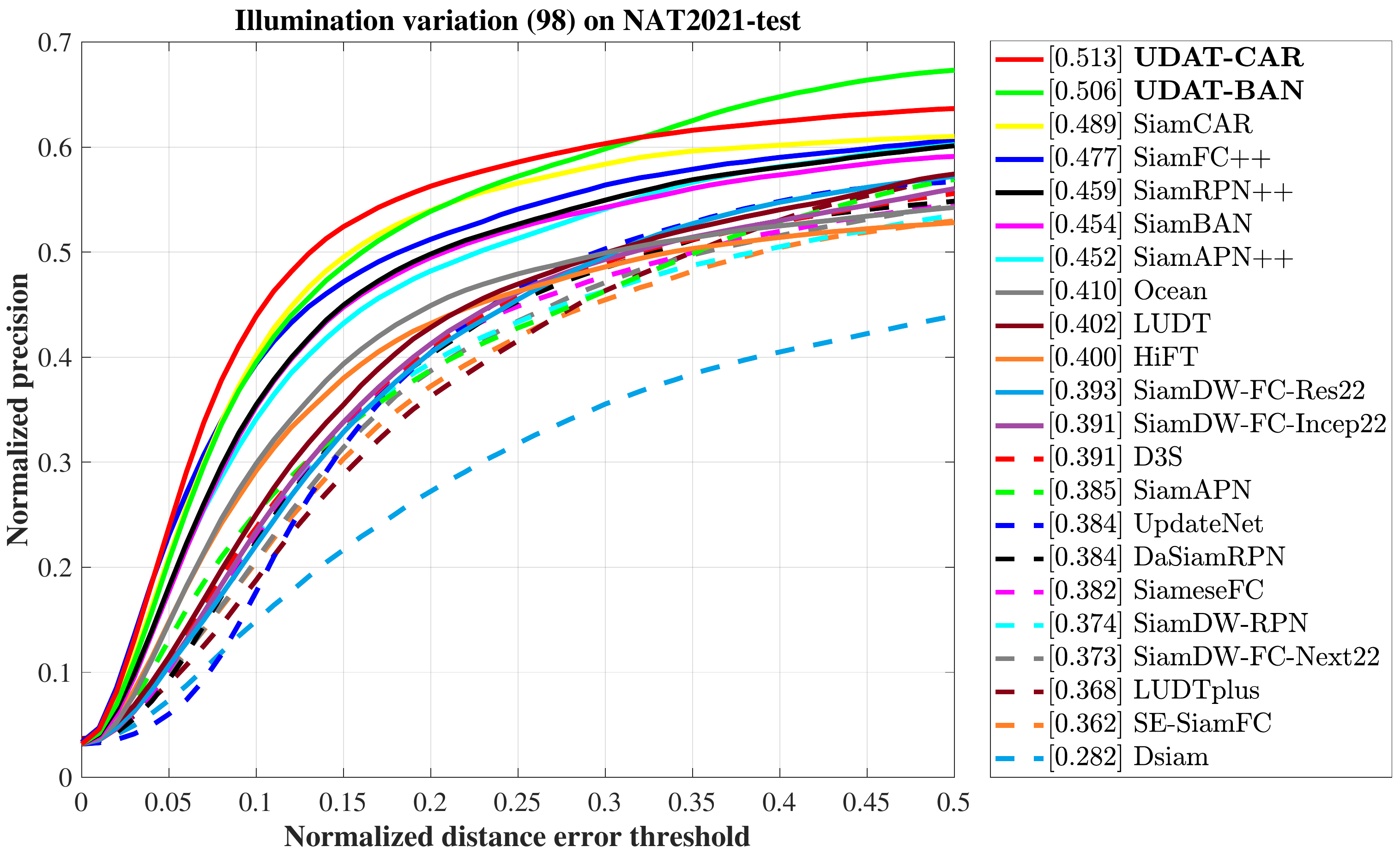}
		\includegraphics[width=0.242\linewidth]{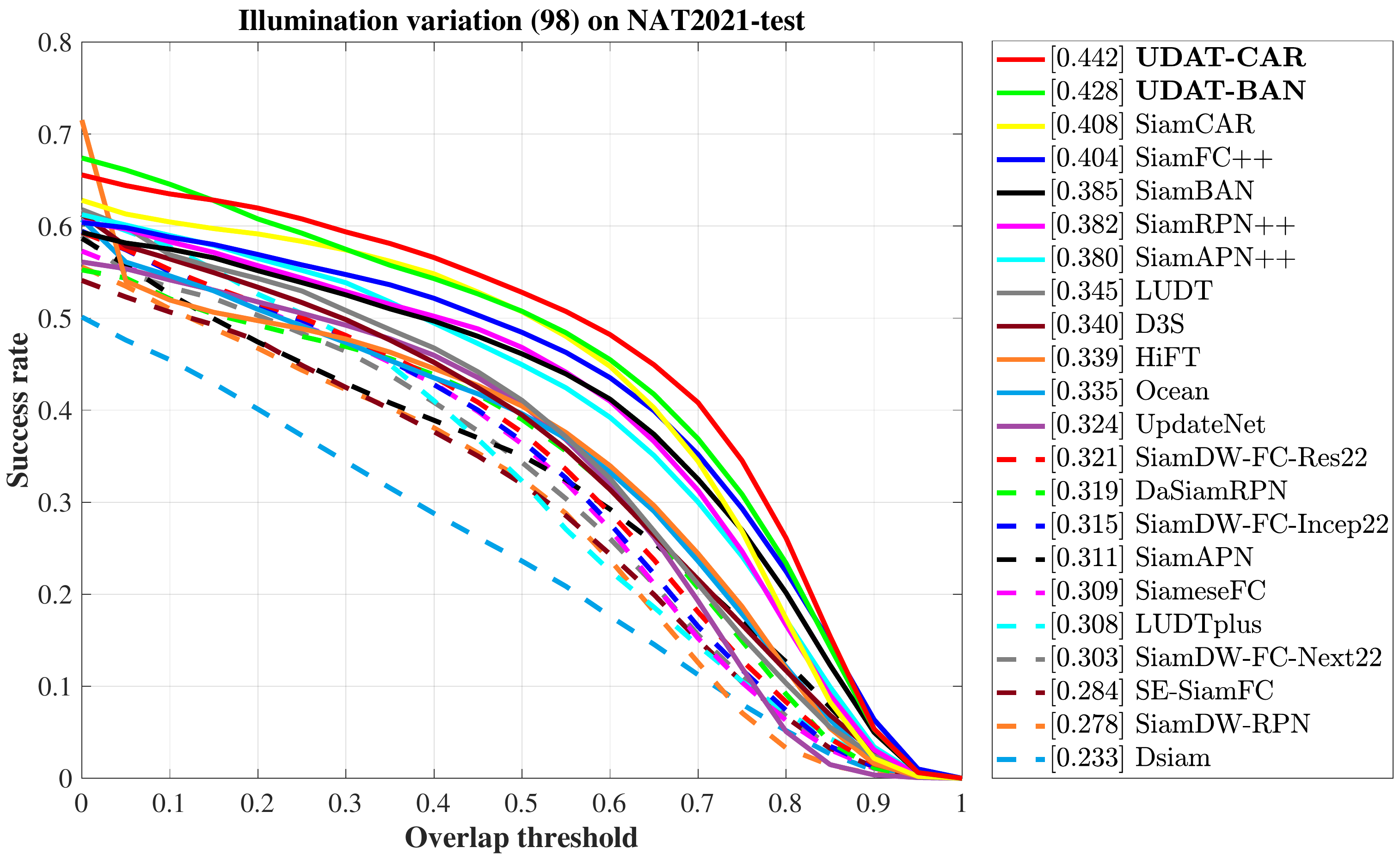}
	}
	\subfloat[Low ambient intensity on NAT2021-$test$.]
	{
		\includegraphics[width=0.242\linewidth]{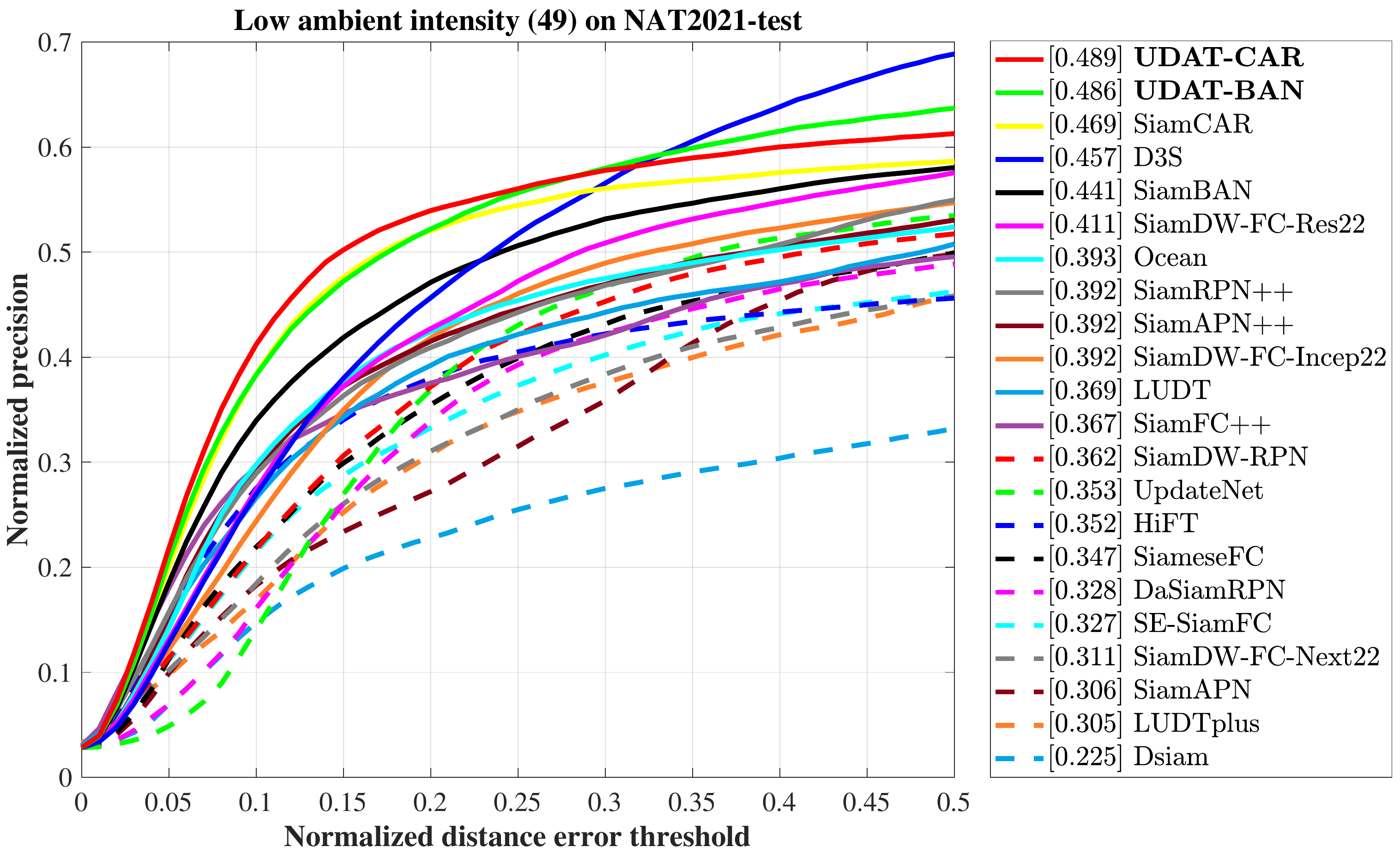}
		\includegraphics[width=0.242\linewidth]{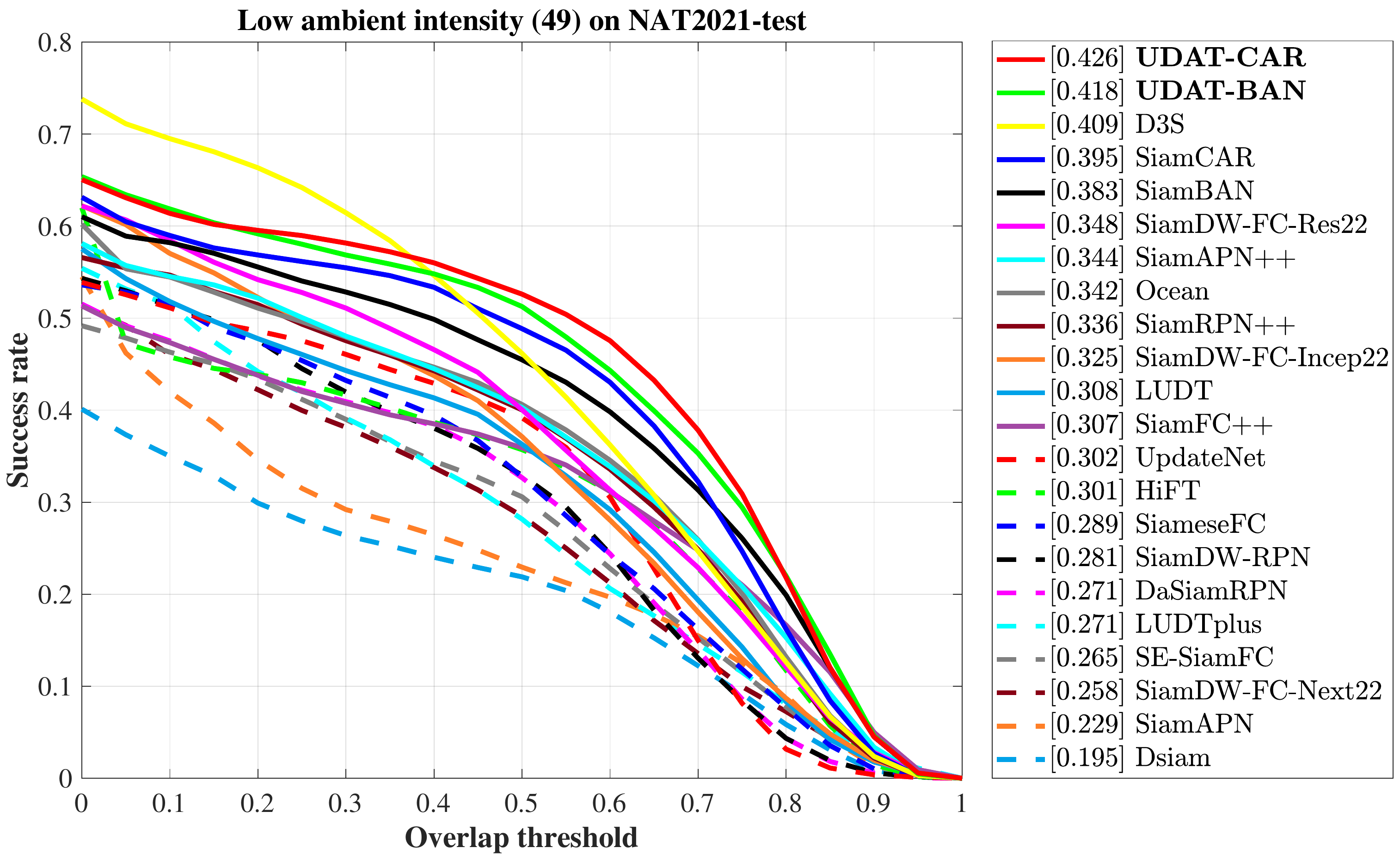}
	}
	
	\subfloat[Illumination variation on UAVDark70.]
	{
		\includegraphics[width=0.242\linewidth]{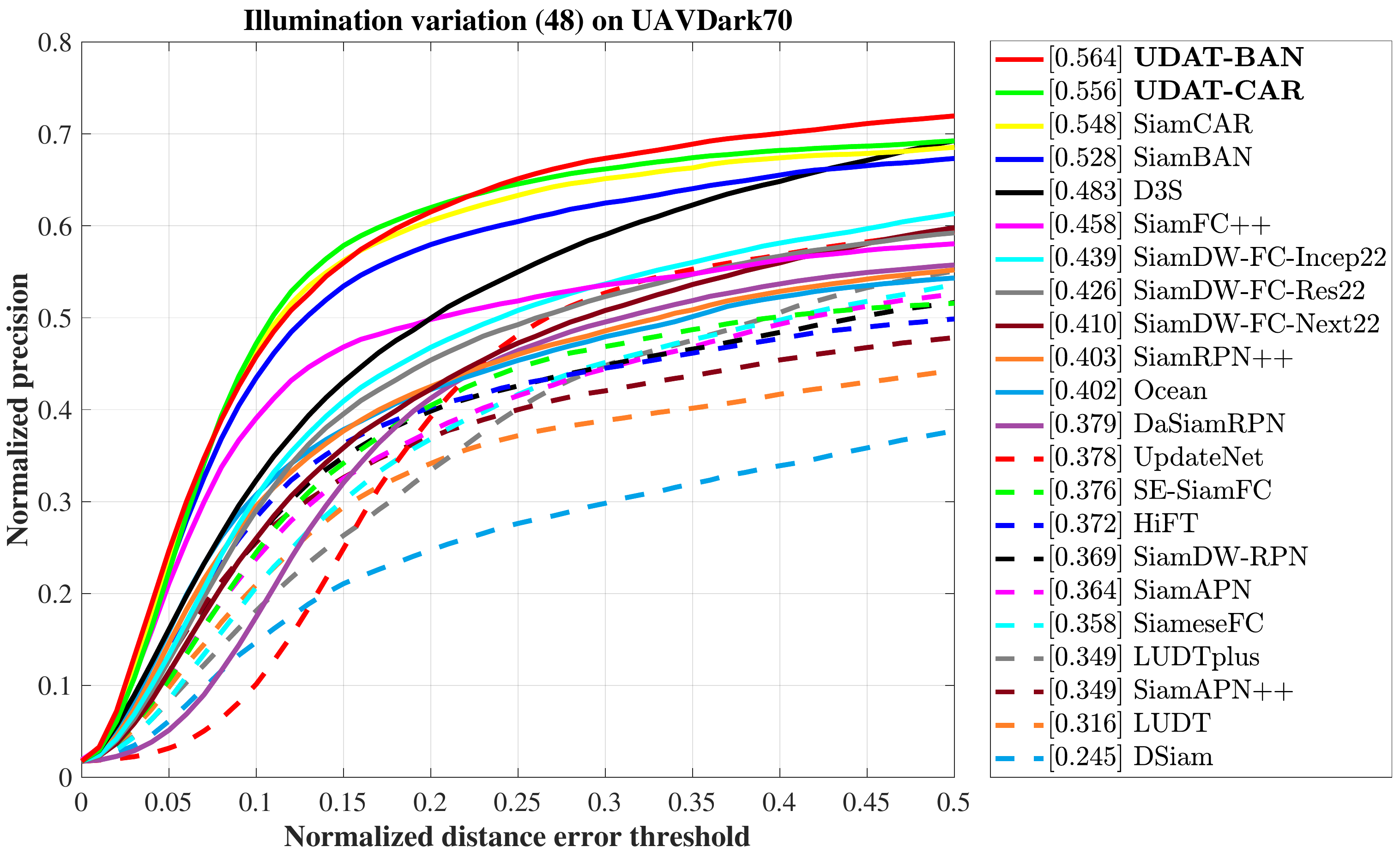}
		\includegraphics[width=0.242\linewidth]{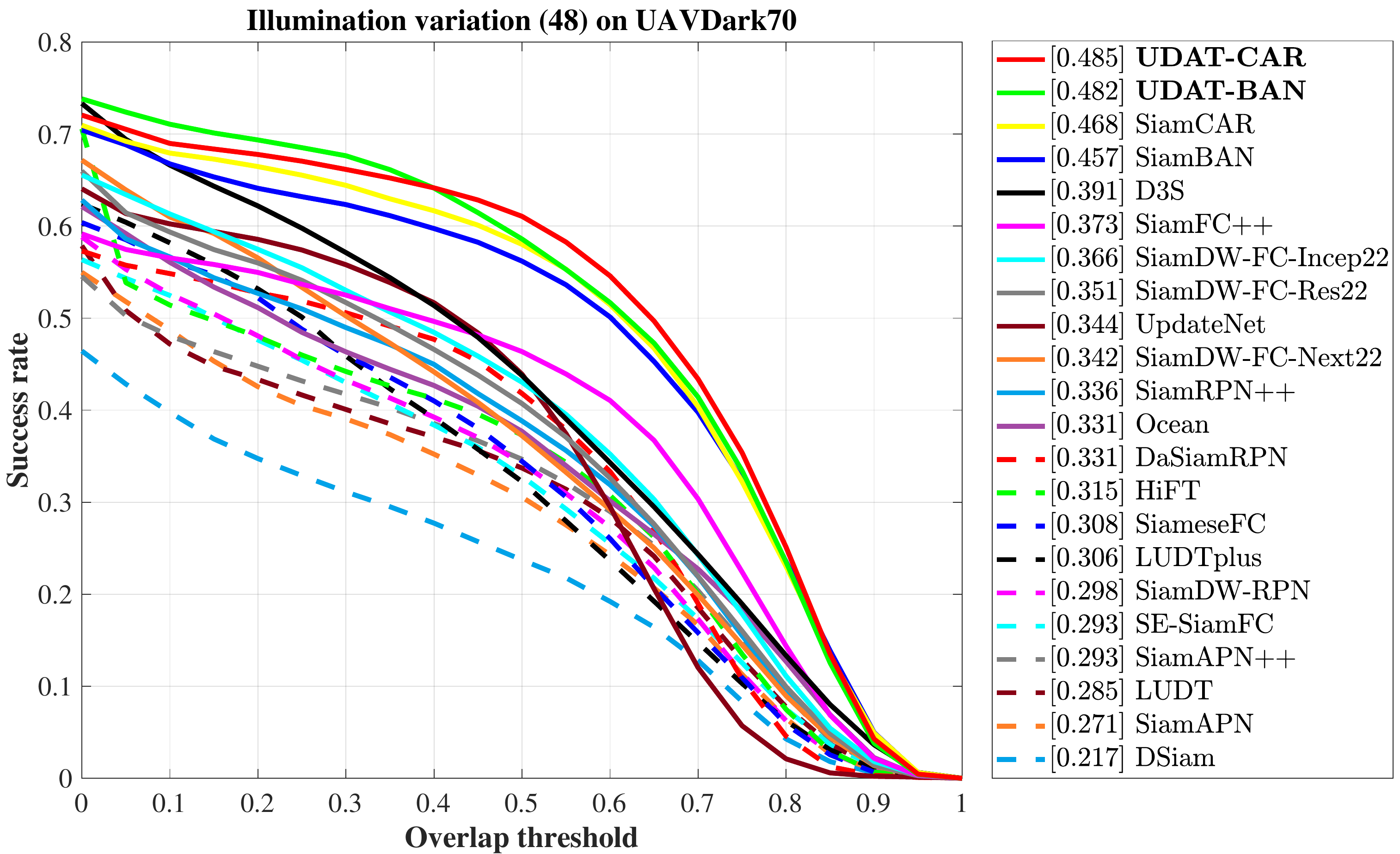}
	}
	\subfloat[Low ambient intensity on UAVDark70.]
	{
		\includegraphics[width=0.242\linewidth]{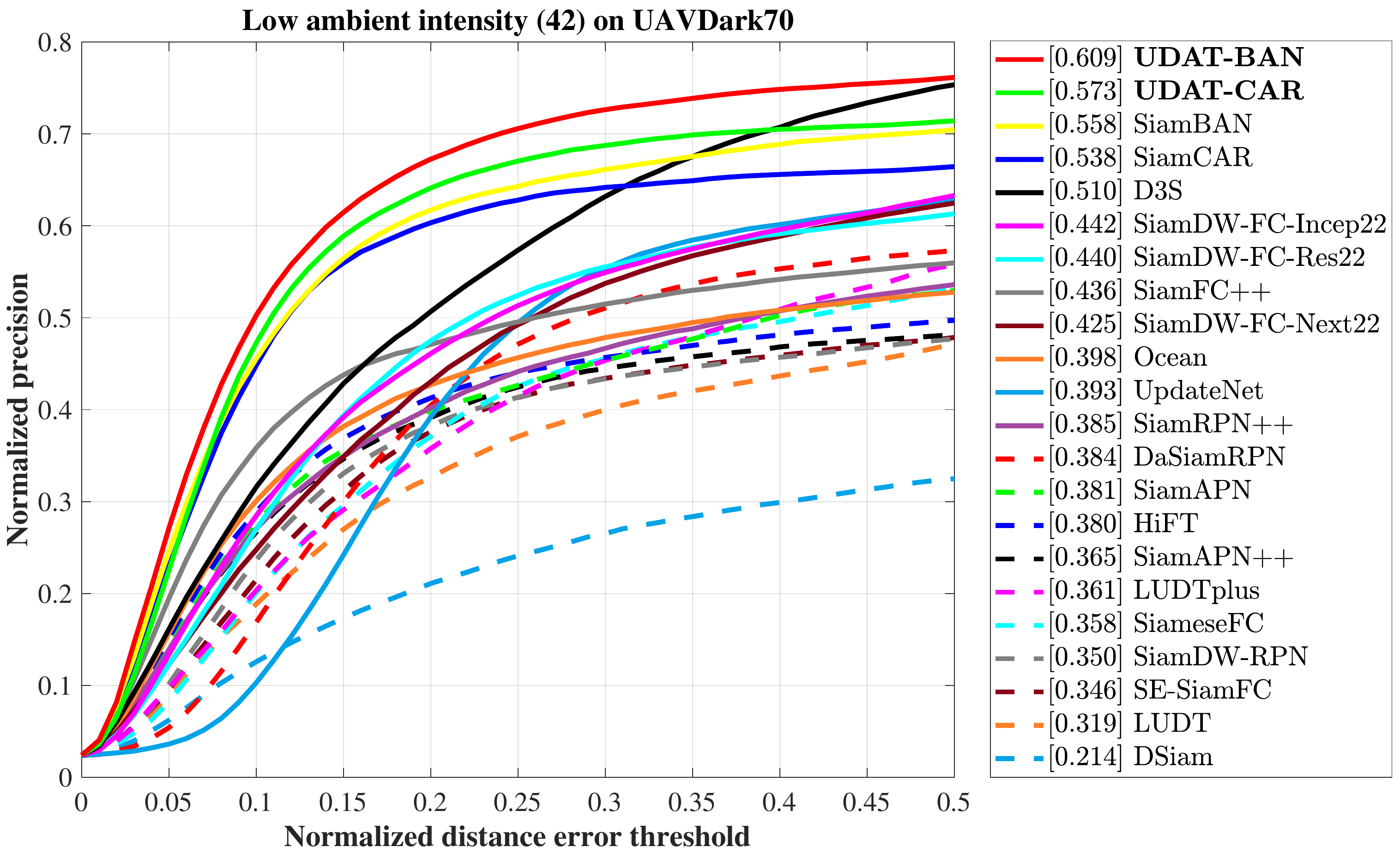}
		\includegraphics[width=0.242\linewidth]{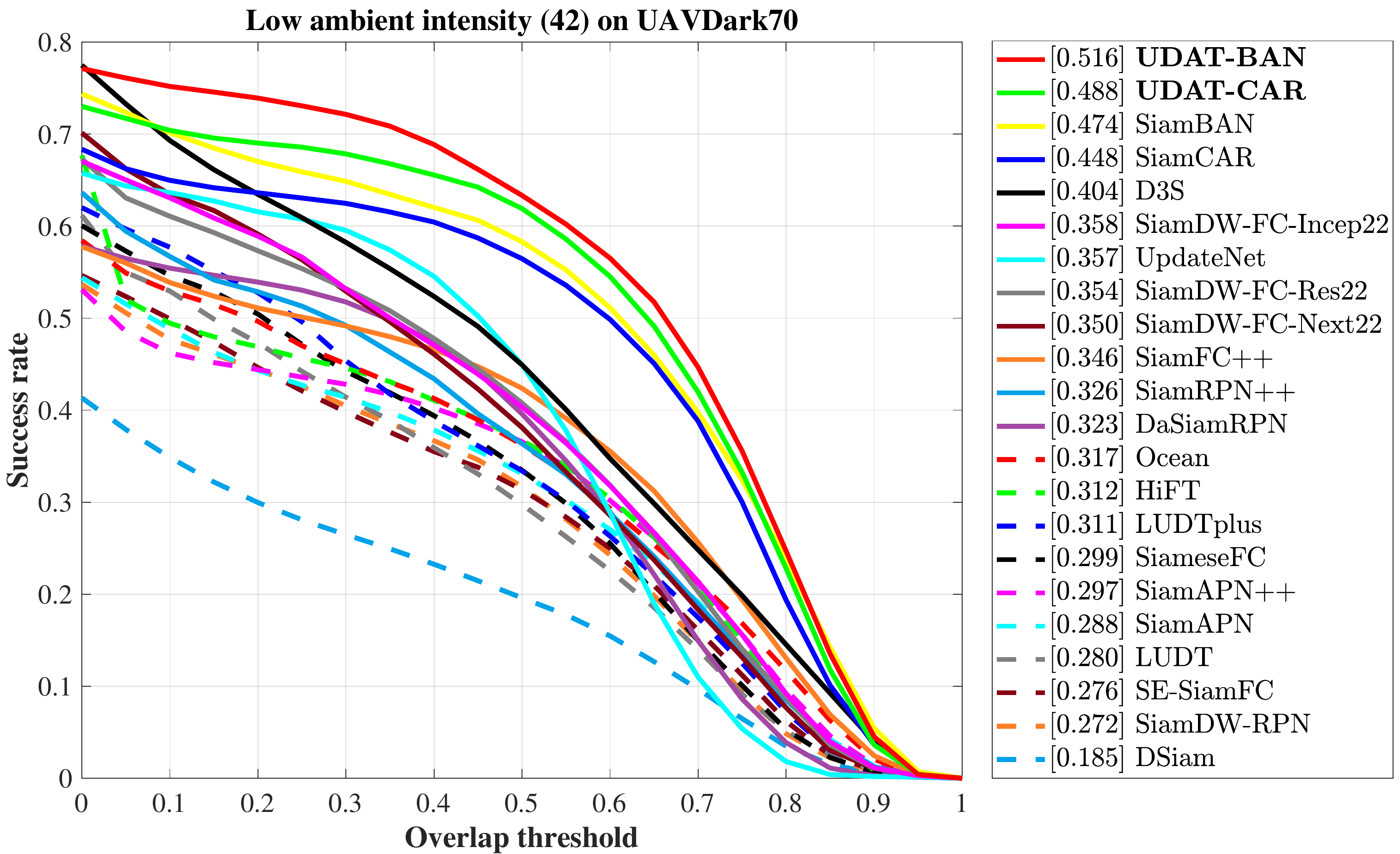}
	}
\setlength{\abovecaptionskip}{3pt}
	\caption
	{
		Normalized precision plots and success plots of illumination-related attributes on NAT2021-$test$ and UAVDark70. 
	}
	\label{fig:att}
		\vspace{-8pt}
\end{figure*}	

% Table generated by Excel2LaTeX from sheet 'Sheet1'
\begin{table*}[!t]
	\centering
	\scriptsize
	\caption{Performance of top-10 trackers on NAT2021-$L\text{-}test$. $\Delta$ represents the percentages of UDAT trackers exceeding the corresponding baselines. The top-2 performance is emphasized with bold font. UDAT trackers yield competitive long-term tracking performance.}
	\vspace{-7pt}
	\renewcommand\tabcolsep{2pt}
	\resizebox{\linewidth}{!}{
		\begin{tabular}{lcccccccccc|cc}
			\toprule
			Trackers & HiFT~\cite{Cao_2021_ICCV} & SiamFC++~\cite{xu2020AAAI} & Ocean~\cite{zhang2020ocean} & SiamRPN++~\cite{Li2019CVPR} & UpdateNet~\cite{Zhang2019ICCV} & D3S~\cite{luke2020CVPR} & SiamBAN~\cite{Chen2020CVPR} & SiamCAR~\cite{Guo2020CVPR} & \bf{UDAT-BAN} & \bf{UDAT-CAR} & \multicolumn{1}{l}{$\Delta_{\rm BAN} (\%)$} & \multicolumn{1}{l}{$\Delta_{\rm CAR} (\%)$} \\
				\midrule
				Prec.  & 0.433  & 0.425  & 0.454  & 0.431  & 0.434  & 0.492  & 0.464  & 0.477  & \textbf{0.496}  & \textbf{0.506}  & 6.94   & 5.99 \\
				Norm. Prec. & 0.316  & 0.344  & 0.370  & 0.342  & 0.314  & 0.364  & 0.366  & 0.375  & \textbf{0.406}  & \textbf{0.413}  & 11.01  & 9.96 \\
				Succ.  & 0.287  & 0.297  & 0.315  & 0.299  & 0.275  & 0.332  & 0.316  & 0.330  & \textbf{0.352}  & \textbf{0.376}  & 11.51  & 14.25 \\
				\bottomrule
		\end{tabular}}
\vspace{-10pt}
	\label{tab:long}%
\end{table*}%

	\subsection{Evaluation results}\label{sec: eval}
	To give an exhaustive analysis of trackers in nighttime aerial tracking and facilitate future research, 20 SOTA trackers~\cite{Guo2020CVPR, Chen2020CVPR, luke2020CVPR, Li2019CVPR, Fu2021ICRA, Cao_2021_IROS, wang2021unsupervised, xu2020AAAI, Zhang2019CVPR, zhang2020ocean, Zhang2019ICCV, Zhu2018ECCV, Bertinetto2016ECCVW,Cao_2021_ICCV,Sosnovik_2021_WACV, Guo2017ICCV} are evaluated on NAT2021-$test$, along with the proposed UDAT. For clarity, two trackers further trained by UDAT are named UDAT-BAN and UDAT-CAR, respectively. Moreover, UAVDark70~\cite{Li2021ICRA} contains 70 nighttime tracking sequences with 66k frames in total, which can also serve as an evaluation benchmark.

	\subsubsection{Overall performance}
	
	\noindent\textbf{NAT2021-$\textit{\textbf{test}}$.} As shown in \cref{fig:all} (a), the proposed UDAT-BAN and UDAT-CAR rank first two places with a large margin compared to their baselines. A performance comparison of UDAT and baseline trackers is reported in \cref{tab:pro}. Specifically, UDAT promotes SiamBAN over \textbf{7}\% on all three metrics. In success rate, UDAT-BAN (0.469) and UDAT-CAR (0.483) raise the original SiamBAN (0.437) and SiamCAR (0.453) by \textbf{7.32}\% and \textbf{6.62}\%, respectively. 

	\noindent\textbf{UAVDark70.} Results in \cref{fig:all} (b) demonstrate that the performance of existing trackers is still unsatisfactory. UDAT trackers raise the performance of their baselines by $\sim$\textbf{4}\%. Note that the data distribution in UAVDark70 is fairly different from that in NAT2021, while UDAT can still bring favorable performance gains, which demonstrate its generalization ability in variant nighttime conditions.
	
	Gains brought by UDAT for different trackers on different benchmarks verify the effectiveness and transferability of the proposed domain adaptation framework.

\subsubsection{Long-term tracking evaluation}	

	As one of the most common scenes in aerial tracking, long-term tracking involves multiple challenging attributes. We further assess trackers on NAT2021-$L\text{-}test$. Top-10 performances are reported in \cref{tab:long}. Results show that UDAT realizes competitive long-term tracking performances, considerably arousing the performance upon baseline trackers.

	\subsubsection{Illumination-oriented evaluation}
	Since the greatest difference between daytime and nighttime tracking is illumination intensity, we perform an in-depth illumination-oriented evaluation for a better analysis of illumination influence on trackers. The results are shown in \cref{fig:att}. Note that we additionally annotate sequences in UAVDark70 with the proposed LAI attribute. The results show that existing trackers suffer from illumination-related attributes. For the IV challenge, the best success rates of existing trackers are 0.408 on NAT2021-$test$ and 0.468 on UAVDark70. Assisted by the proposed domain adaptive training, UDAT-CAR realizes a success rate of 0.442 and 0.485, respectively, which fairly improve the existing best performance. As for LAI, UDAT-BAN raises the normalized precision of its baseline SiamBAN by over \textbf{9}\% on both benchmarks. From the comparison, we can see that trackers' illumination-related performance remains a large room for improvement and the adoption of domain adaptation in adverse illumination scenes is effective and crucial.

	%%%%%%%%%%%%%%%%%%%%%%%%%%%%%%

	\subsubsection{Visualization}
\begin{figure}[!t]	
	\centering
	\includegraphics[width=0.99\linewidth]{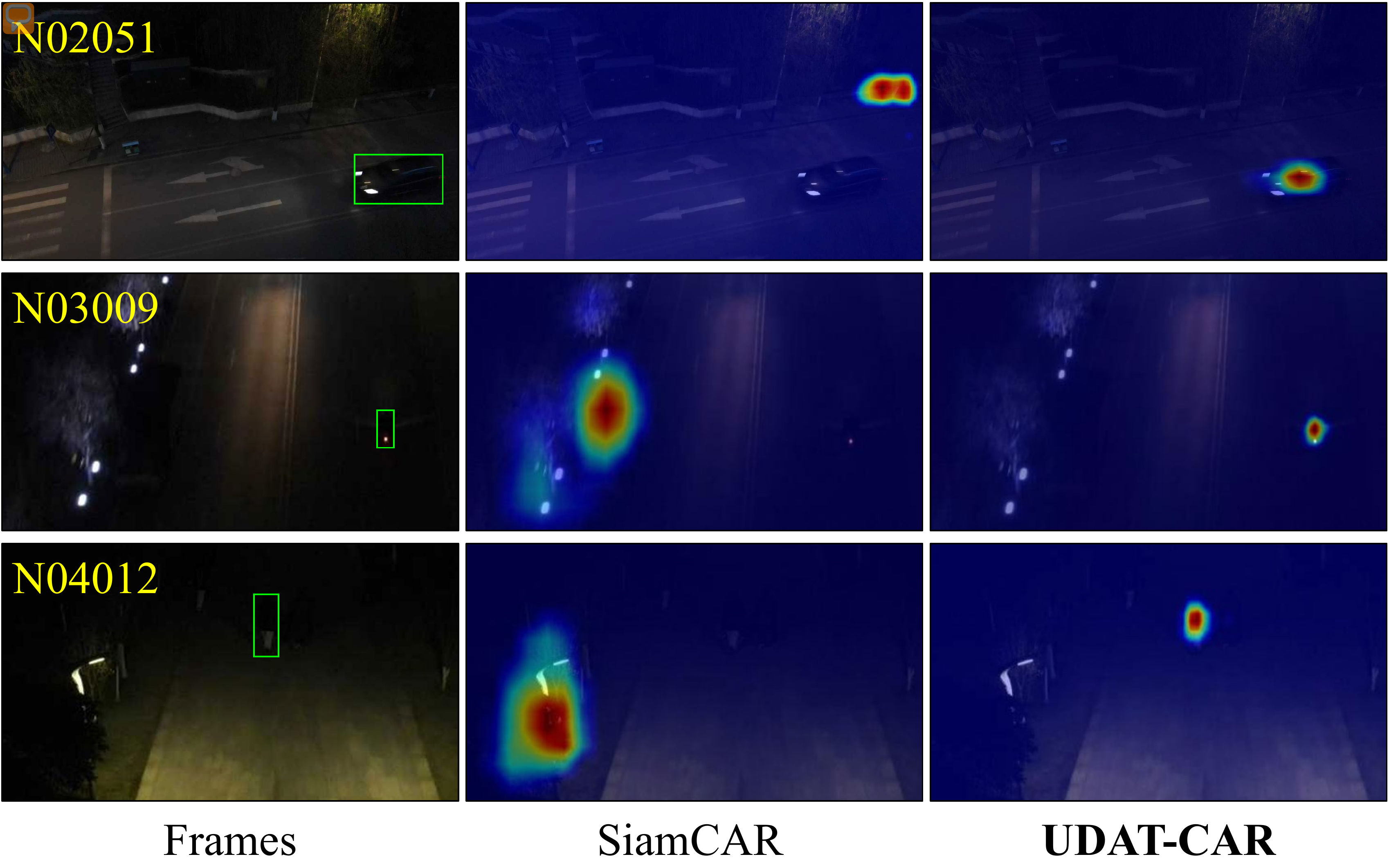}
	\setlength{\abovecaptionskip}{5pt}
	\caption
	{Visual comparison of confidence maps generated by the baseline and the proposed UDAT. Target objects are marked by \NoTwo{green} boxes. The baseline struggles to extract discriminable features in dim light. UDAT substantially raises the perception ability of baseline in adverse illuminance.
	}  
\vspace{-10pt}
	\label{fig: attention}
\end{figure}

	As shown in \cref{fig: attention}, we visualized some confidence maps of UDAT and its baseline using Grad-Cam~\cite{Selvaraju2017ICCV}. The baseline model fails to concentrate on objects in adverse illuminance, while UDAT substantially enhances the baseline's nighttime perception ability, thus yielding satisfying nighttime tracking performance.
	
	\subsubsection{Source domain evaluation}
	Apart from favorable performance at nighttime, we expect that trackers do not suffer degradation at the source domain during adaptation. Evaluation on a daytime tracking benchmark UAV123~\cite{Mueller2016ECCV} is shown in~\cref{tab:source}. The results show that UDAT brings slight performance fluctuation within 2\% in success rate and 0.5\% in precision.
	
	\begin{table}[!b]
		\centering
		\scriptsize
		\caption{Evaluation on the source domain. The results show the adaptation only brings slight performance fluctuation on the source domain.}
		\vspace{-6pt}
			\begin{tabular}{l|cc|cc}
				\toprule
				Trackers & SiamBAN & UDAT-BAN & SiamCAR & UDAT-CAR \\
				\midrule
				Succ.  & 0.603  & 0.591$_{1.96\%\downarrow}$  & 0.601  & 0.592$_{1.58\%\downarrow}$ \\
				Prec.  & 0.788  & 0.784$_{0.52\%\downarrow}$  & 0.793  & 0.793$_{0.04\%\downarrow}$ \\
				\bottomrule
			\end{tabular}%
		\vspace{-3pt}
		\label{tab:source}%
	\end{table}%
% Table generated by Excel2LaTeX from sheet 'Sheet1'
\begin{table}[!b]
	\centering
	\scriptsize
	\caption{Empirical Study of the proposed UDAT on NAT2021-$test$. \texttt{DA}, \texttt{OD}, and \texttt{BL} denote domain adaptive training, object discovery preprocessing, and bridging layer, respectively.}
	\vspace{-6pt}
	\renewcommand\tabcolsep{7.5pt}
	\begin{tabular}{cccccc}
		\toprule
		\texttt{DA}     & \texttt{OD}     & \texttt{BL}     & Prec.  & Norm. Prec. & Succ. \\
		\midrule
		&        &        & 0.663  & 0.542  & 0.453 \\
		\midrule
		\checkmark &        &        & 0.662$_{0.19\%\downarrow}$  & 0.540$_{0.33\%\downarrow}$  & 0.459$_{1.33\%\uparrow}$ \\
		\checkmark &        & \checkmark & 0.664$_{0.16\%\uparrow}$  & 0.547$_{1.04\%\uparrow}$  & 0.464$_{2.45\%\uparrow}$ \\
		\checkmark & \checkmark &        & 0.676$_{1.95\%\uparrow}$  & 0.549$_{1.42\%\uparrow}$  & 0.467$_{3.24\%\uparrow}$ \\
		\midrule
		\checkmark & \checkmark & \checkmark & \textbf{0.687}$_{3.62\%\uparrow}$  & \textbf{0.564}$_{4.17\%\uparrow}$  & \textbf{0.483}$_{6.82\%\uparrow}$ \\
		\bottomrule
	\end{tabular}
%	\vspace{-10pt}
	\label{tab:abla}%
\end{table}%

	\subsection{Empirical study}
	To demonstrate the effectiveness of proposed modules, \ie, domain adaptive training (\texttt{DA}), object discovery preprocessing (\texttt{OD}), and bridging layer (\texttt{BL}), this subsection provides empirical studies of UDAT. 
	Concretely, we first ablate \texttt{BL} and substitute \texttt{OD} with random cropping to adopt naive \texttt{DA} on the baseline tracker. The results on the second row of \cref{tab:abla} show that \texttt{DA} slightly promotes nighttime tracking, with a slight upgrade in success rate. However, adopting random cropping as preprocessing leads to abundant meaningless training samples, the model therefore can hardly learn the data distribution on the target domain. In that case, further activation of \texttt{BL} only makes a limited difference. As shown in the fourth row of \cref{tab:abla}, when employing \texttt{OD} instead of random cropping, performance on the target domain obtains a 3.24\% boost in success rate, which verifies the effectiveness of the proposed saliency detection-based data preprocessing. Further, \texttt{BL} doubles the promotion brought by \texttt{OD}, complete UDAT realizes a precision of 0.687 and a success rate of 0.483, achieving favorable nighttime tracking performance. The results verify that the proposed bridging layer fairly enables the tracker to generate discriminative features from nighttime images.

	\section{Conclusion}
	In this work, a simple but effective unsupervised domain adaptive tracking framework, namely UDAT, is proposed for nighttime aerial tracking. In our UDAT, an object discovery strategy is introduced for unlabelled data preprocessing. The Transformer bridging layer is adopted to narrow the gap of image features between daytime and nighttime. Optimized through adversarial learning with a Transformer discriminator, the learned model substantially improves nighttime tracking performance upon SOTA approaches. We also construct NAT2021, a pioneering benchmark for unsupervised domain adaptive nighttime tracking. Detailed evaluation on nighttime tracking benchmarks shows the effectiveness and domain adaptability of UDAT. The limitation of this work lies in the absence of pseudo supervision in the target domain. Future work will focus on reliable pseudo supervision, with which we believe the performance of nighttime tracking can be further improved. To sum up, we are convinced that the UDAT framework along with the NAT2021 benchmark can facilitate research on visual tracking at nighttime and in other adverse conditions.
	
	\noindent\textbf{Acknowledgement:} This work was supported in part by the National Natural Science Foundation of China under Grant 62173249 and in part by the Natural Science Foundation of Shanghai under Grant 20ZR1460100.
	
	%%%%%%%%% REFERENCES
	{\small
		\bibliographystyle{ieee_fullname}
		\bibliography{egbib}
	}
	
\end{document}